\documentclass[12pt,twoside]{article}
\usepackage{graphicx,hyperref}
  \newdimen\paravsp  \paravsp=1.3ex
\def\keyword#1{\centerline{\bf\small
Keywords}\begin{quote}\small #1 \par\end{quote}\vskip 1ex}
\usepackage{latexsym,amsmath,amssymb}
\usepackage{wrapfig}

\newdimen\paravsp  \paravsp=1.3ex
\def\paradot#1{\vspace{\paravsp plus 0.5\paravsp minus 0.5\paravsp}\noindent{\bf\boldmath{#1.}}}
\def\length{\text{length}}
\def\qmbox#1{{\quad\mbox{#1}\quad}}
\def\Loss{\text{Loss}}
\def\E{\mathbb{E}}

\def\frs#1#2{{^{#1}\!/\!_{#2}}}
\def\M{{\cal M}}
\def\X{{\cal X}}
\def\t{\theta}
\def\Km{K} 
\def\limpl{\Rightarrow} 

\begin{document}

\title{\vspace{-3ex}
\vskip 2mm\bf\Large\hrule height5pt \vskip 4mm
A Philosophical Treatise of Universal Induction
\vskip 4mm \hrule height2pt}
\author{{\bf Samuel Rathmanner} and {\bf Marcus Hutter}\\[3mm]
\normalsize Research School of Computer Science \\[-0.5ex]
\normalsize Australian National University}
\date{25 May 2011}
\maketitle

\begin{abstract}
Understanding inductive reasoning is a problem that
has engaged mankind for thousands of years. This problem is
relevant to a wide range of fields and is integral to the
philosophy of science. It has been tackled by many great minds
ranging from philosophers to scientists to mathematicians, and
more recently computer scientists. In this article we argue the
case for Solomonoff Induction, a formal inductive framework
which combines algorithmic information theory with the Bayesian
framework. Although it achieves excellent theoretical results
and is based on solid philosophical foundations, the requisite
technical knowledge necessary for understanding this framework
has caused it to remain largely unknown and unappreciated in
the wider scientific community. The main contribution of this
article is to convey Solomonoff induction and its related
concepts in a generally accessible form with the aim of
bridging this current technical gap. In the process we examine
the major historical contributions that have led to the
formulation of Solomonoff Induction as well as criticisms of
Solomonoff and induction in general. In particular we examine
how Solomonoff induction addresses many issues that have
plagued other inductive systems, such as the black ravens
paradox and the confirmation problem, and compare this approach
with other recent approaches.
\end{abstract}

\keyword{
sequence prediction; %
inductive inference; %
Bayes rule; %
Solomonoff prior; %
Kolmogorov complexity; %
Occam's razor; %
philosophical issues; %
confirmation theory; %
Black raven paradox. %
}

\begin{quote}\it
This article is dedicated to Ray Solomonoff (1926--2009), \\
the discoverer and inventor of Universal Induction.
\hfill\smash{\raisebox{-12ex}{\includegraphics[width=3.0cm]{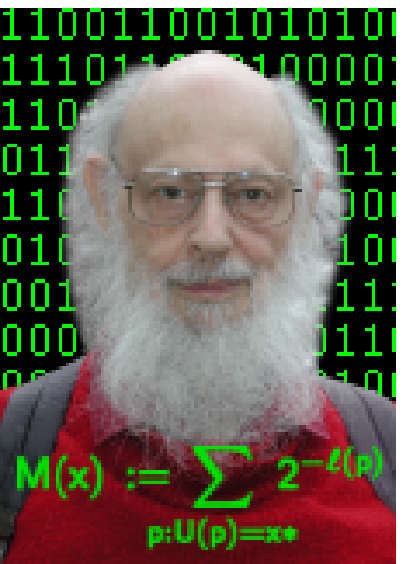}\hspace{0ex}}}
\end{quote}

\newpage
{\footnotesize\parskip=-1ex\tableofcontents}

\newpage
\section{Introduction}\label{cha:intro}

According to Aristotle, it is our ability to reason which sets
humans apart from the rest of the animal kingdom. The
understanding and manipulation of our environment that has
made us so successful has only been possible through this
unique ability. Reasoning underpins every human advancement and
is used on a daily basis even if only trivially. Yet
surprisingly, although reasoning is fundamental to the
functioning and evolution of our species, we have had great
difficulty in producing a satisfactory explanation of the
mechanism that governs a large portion of this reasoning;
namely inductive reasoning.

The difficulty of any attempt at unraveling the inner workings
of the human mind should be appreciated. Some have even argued
that a complete understanding of the human mind is beyond the
capabilities of the human mind \cite{McGinn:89}. Understanding
how we reason is however an area in which significant progress
has been made.

Reasoning is often broken down into two broad categories. Firstly
there is {\em deductive reasoning} which can be thought of as the
process of drawing logically valid conclusions from some assumed or
given premise. Deductive reasoning is the type of reasoning used in
mathematical proofs or when dealing with formal systems. Although
this type of reasoning is obviously necessary it is not always adequate.

When reasoning about our world we often want to make
predictions that involve estimations and generalizations. For
this we use {\em inductive reasoning}. Inductive reasoning can
be thought of as drawing the `best' conclusions from a set of
observations. Unfortunately these observations are almost
always incomplete in some sense and therefore we can never be
certain of the conclusions we make. This process is analogous
to the scientific process in general. In science, rules and
models are found by generalizing patterns observed locally.
These models are then used to understand and predict our
environment which in turn allows us to benefit, usually with
great success. But like inductive inference, scientific
hypotheses can never be completely validated, so we can never
know whether they are true for certain. The difference between
reasoning inductively or deductively can also be simply thought
of as the difference between reasoning about the known or the
unknown.

Philosophically speaking the fundamental goal of inductive
reasoning is to gain a deeper awareness of how we should
maintain rational beliefs and predictions in the face of
uncertainty and the unknown observations or problems of the
future. In some sense a history of inductive reasoning is a
history of questioning and attempting to understand our own
thought processes. As early as 300BC Epicurus was interested
in how we judge competing theories for some given observations
\cite{Asmis:84}. This led him to postulate his principle of
multiple explanations which stated that we should never
disregard a consistent theory. William of Occam countered this
with his with well-known `Occam's razor' which advised that all
but the simplest theory consistent with the observations should
be discarded \cite{Ockham:90}. Hume later stated the problem of
induction explicitly for the first time {\em ``What is the
foundation of all conclusions from experience?''} \cite{Hume:1739}.
He also set about questioning the validity of such conclusions.
Hume's problem led Bayes and Laplace to make the first attempts
at formalizing inductive inference which has become the basis
for Bayesianism. This is a school of thought that requires
making an explicit choice in the class of explanations
considered and our prior belief in each of them. Bayesianism
has, in turn, fueled further attempts at formalizing induction.

To say that the history of induction has been contentious is an
understatement \cite{McGrayne:11}. There have been many
attempts at formalizing inductive reasoning \cite{Gabbay:11}
that address specific situations and satisfy many of the
intuitive properties we expect of induction. Unfortunately most
of these attempts have serious flaws or are not general enough.
Many of the results regarding induction have been controversial
and highly contested, which is not particularly surprising. By
its very nature induction deals with uncertainty, subjectivity
and challenging philosophical questions, and is therefore
highly prone to discussion and debate. Even if a result is
formally sound, its philosophical assumptions and applicability
in a range of situations are often questioned.

In 1964 Ray Solomonoff published the paper {\em ``A Formal
Theory of Inductive Inference''} \cite{Solomonoff:64}. In this
paper he proposed a universal method of inductive inference
which employs the Bayesian framework and his newly created
theory of algorithmic probability. This method of Solomonoff
induction appears to address the issues that plagued previous
attempts at formalizing induction and has many promising
properties and results. Solomonoff induction and related
concepts are the central focus of this article.

The formalization of Solomonoff induction makes use of concepts
and results from computer science, statistics, information
theory, and philosophy. It is interesting that the development
of a rigorous formalization of induction, which is fundamental
to almost all scientific inquiry, is a highly
multi-disciplinary undertaking, drawing from these various
areas. Unfortunately this means that a high level of technical
knowledge from these various disciplines is necessary to fully
understand the technical content of Solomonoff induction. This
has restricted a deep understanding of the concept to a fairly
small proportion of academia which has hindered its discussion
and hence progress.

In this article we attempt to bridge this gap by conveying the
relevant material in a much more accessible form. In particular
we have expanded the material in the dense 2007 paper {\em ``On
Universal Prediction and  Bayesian Confirmation''}
\cite{Hutter:07uspx} which argues that Solomonoff induction
essentially solves the induction problem. In addition to
providing intuition behind the overall setup and the main
results we also examine the philosophical motivations and
implications of Solomonoff's framework.

We have attempted to write this article in such a way that the
main progression of ideas can be followed with minimal
mathematical background. However, in order to clearly
communicate concrete results and to provide a more complete
description to the capable reader, some technical explanation
is necessary.

By making this knowledge more accessible we hope to promote
discussion and awareness of these important ideas within a much
wider audience. Every major contribution to the foundations of
inductive reasoning has been a contribution to understanding
rational thought. Occam explicitly stated our natural
disposition towards simplicity and elegance. Bayes inspired the
school of Bayesianism which has made us much more aware of the
mechanics behind our belief system. Now, through Solomonoff, it
can be argued that the problem of formalizing optimal inductive
inference is solved.

Being able to precisely formulate the process of (universal)
inductive inference is also hugely significant for general
artificial intelligence. Obviously reasoning is synonymous with
intelligence, but true intelligence is a theory of how to act on
the conclusions we make through reasoning. It may be argued
that optimal intelligence is nothing more than optimal
inductive inference combined with optimal decision
making. Since Solomonoff provides optimal inductive inference
and decision theory solves the problem of choosing optimal
actions, they can therefore be combined to produce intelligence.
This is the approach taken by the second author in developing
the AIXI model which will be discussed only briefly.

\subsection{Overview of Article}

Here we will give a brief summary of the contents of this
article.

{\em Section~\ref{cha:context}} looks at the broader context of
(universal) induction. At first we contrast it with deduction.
We then argue that any inductive inference problem can be
considered or converted to a sequence prediction problem. This
gives justification for focusing on sequence prediction
throughout this article. We also examine the relation between
Solomonoff induction and other recent approaches to induction.
In particular how Solomonoff induction addresses non-monotonic
reasoning and why it appears to contradict the conclusion of
no-free-lunch theorems.

{\em Section~\ref{cha:prob}} covers probability theory and the
philosophy behind the varying schools of thought that exist.
We explain why the subjective interpretation of probability is
the most relevant for universal induction and why it is valid.
In particular we explain why the belief system of any rational
agent must obey the standard axioms of probability.

Applying the axioms of probability to make effective
predictions results in the Bayesian framework which is
discussed in depth in {\em Section~\ref{cha:bayes}}. We look at what
it means to be a Bayesian; why models, environments and
hypotheses all express the same concept; and finally we explain
the mechanics of the Bayesian framework. This includes looking
at convergence results and how it can be used to make optimal
Bayesian decisions. We also look briefly at continuous model
classes and at making reasonable choices for the model class.

{\em Section~\ref{cha:history}} gives an overview of some of the
major historical contributions to inductive reasoning. This
includes the fundamental ideas of Epicurus's principle of
multiple explanations and Occam's razor. We also discuss
briefly the criticisms raised by inductive skeptics such as
Empiricus and Hume. Laplace's contribution is then examined.
This includes the derivation of his famous rule of succession
which illustrates how the Bayesian framework can be applied.
This rule also illustrates the confirmation problem that
has plagued many attempts at formalizing induction. The cause
of this problem is examined and we show that a recent claim by
Maher \cite{Maher:04}, that the confirmation problem can be
solved using the axioms of probability alone,
is clearly unsatisfactory. One of the most difficult problems
with confirmation theory is the black ravens paradox. We
explain why this counter-intuitive result arises and the
desired solution.

In order to understand the concept of Kolmogorov complexity
which is integral to Solomonoff induction, it is necessary to
briefly examine the fundamental concept of computability and
the closely related Turing machine. The introduction of the
theoretical Turing machine is in some ways considered the birth
of computer science. We look at the basic idea of what a Turing
machine is and how, through Solomonoff, it became an integral
part of universal induction. The measure of complexity that
Kolmogorov developed is a major part of algorithmic information
theory. We examine the intuition behind it as well as some
relevant properties.

In {\em Section~\ref{cha:prior}} we discuss reasonable approaches to
making a rational choice of prior as well as desirable
properties for a prior to have. The universal prior involves
the use of Kolmogorov complexity which we argue is highly
intuitive and does justice to both Occam and Epicurus.

At this point, having introduced all the necessary concepts,
{\em Section~\ref{cha:solomonoff}} gives an explanation of
Solomonoff's universal predictor. We examine two alternate
representations of this universal predictor and the
relationship between them. We also look at how this predictor
deals with the problem of old evidence, the confirmation problem
and the black ravens paradox.

{\em Section~\ref{cha:bounds}} discusses several bounds for this
universal predictor, which demonstrate that it performs
excellently in general. In particular we present total,
instantaneous, and future error bounds.

{\em Section~\ref{cha:app}} looks at the value of Solomonoff
induction as a gold standard and how it may be approximated and
applied in practice. We mention a number of approximations and
applications of either Solomonoff or the closely related
Kolmogorov complexity. The extension of Solomonoff to universal
artificial intelligence is also briefly covered.

{\em Section~\ref{cha:discussion}} gives a brief discussion of some
of the issues concerning Solomonoff induction as well as a
review of the pro's and con's, and concludes.

\section{Broader Context}\label{cha:context}

The work done on the problem of induction, both philosophically
and formally, has been both vast and varied.
In this article the focus is on using inductive inference for
making effective decisions. From this perspective, having an
effective method of prediction is sufficient. It is for this
reason that this article focuses primarily on sequence
prediction, rather than inference in the narrow sense of
learning a general model from specific data.
Even concept learning, classification, and regression problems
can be rephrased as sequence prediction problems.
After having clarified these relationships, we briefly look at
Solomonoff induction in the context of some of the most
significant concepts in recent discussions of inductive
reasoning \cite{Gabbay:11} such as Bayesian learning versus
prediction with expert advice, no-free-lunch theorems versus
Occam's razor, and non-monotonic reasoning.

\subsection{Induction versus Deduction}

There are various informal definitions of inductive inference. It can be
thought of as the process of deriving general conclusions from
specific observation instances. This is sort of dual to
deductive inference, which can be thought of as the process of
deducing specific results from general axioms. These
characterizations may be a bit narrow and misleading, since
induction and deduction also parallel each other in certain
respects.

The default system for deductive reasoning is ``classical''
(first-order predicate) logic. The Hilbert calculus starts with
a handful of logical axiom schemes and only needs modus ponens
as inference rule. Together with some domain-specific
non-logical axioms, this results in ``theorems''. If some
real-world situation (data, facts, observation) satisfies the
conditions in a theorem, the theorem can be applied to derive
some conclusions about the real world. The axioms in
Zermelo-Fraenkel set theory are universal in the sense that all
of mathematics can be formulated within it.

Compare this to the (arguably) default system for inductive
reasoning based on (real-valued) probabilities: The Bayesian
``calculus'' starts with the Kolmogorov axioms of probability,
and only needs Bayes rule for inference. Given some
domain-specific prior and new data=facts=observations, this
results in ``posterior'' degrees of belief in various
hypotheses about the world. Solomonoff's prior is universal in
the sense that it can deal with arbitrary inductive inference
problems.
Hypotheses play the role of logical expressions, probability
$P(X)=0/1$ corresponds to $X$ being false/true, and $0<P(X)<1$
to $X$ being true in some models but false in others.
The general correspondence is depicted in the following table:

\begin{center}
\begin{tabular}{l|rcl} \hline
                      & \bf Induction            & \boldmath$\Leftrightarrow$ & \bf Deduction \\ \hline
  Type of inference:  & generalization/prediction& $\Leftrightarrow$          & specialization/derivation \\
  Framework:          & probability axioms       & $\widehat=$                & logical axioms \\ 
  Assumptions:        & prior                    & $\widehat=$                & non-logical axioms \\   
  Inference rule:     & Bayes rule               & $\widehat=$                & modus ponens \\
  Results:            & posterior                & $\widehat=$                & theorems \\
  Universal scheme:   & Solomonoff probability   & $\widehat=$                & Zermelo-Fraenkel set theory \\
  Universal inference:& universal induction      & $\widehat=$                & universal theorem prover \\ \hline
\end{tabular}
\end{center}

\subsection{Prediction versus Induction}

The above characterization of inductive inference as the
process of going from specific to general was somewhat narrow.
Induction can also be understood more broadly to include the
process of drawing conclusions about some given data, or even
as the process of predicting the future.
Any inductive reasoning we do must be based on some data or
evidence which can be regarded as a history. From this data we
then make inferences, see patterns or rules or draw conclusions
about the governing process. We are not really interested in
what this tells us about the already observed data since this
is in the past and therefore static and inconsequential to
future decisions. Rather we care about what we are able to
infer about future observations since this is what allows
us to make beneficial decisions. In other words we want to
predict the natural continuation of our given history of
observations. Note that `future observations' can also refer to
past but (so far) unknown historical events that are only
revealed in the future.

From this general perspective, the scientific method can be
seen as a specific case of inductive reasoning. In science we
make models to explain some past data or observation history
and these models are then used to help us make accurate
predictions. As humans we find these models satisfying as we
like to have a clear understanding of what is happening, but
models are often overturned or revised in the future. Also,
from a purely utilitarian point of view, all that matters is
our ability to make effective decisions and hence only the
predictions and not the models themselves are of importance.
This is reminiscent of but not quite as strong as the famous
quote by George Box that ``Essentially, all models are wrong,
but some are useful''.

We look now at some specific examples of how general induction
problems can be rephrased as prediction problems.

\subsection{Prediction, Concept Learning, Classification, Regression}

In many cases the formulation is straight forward.
For example, problems such as
``what will the weather be like tomorrow?'', ``what will the
stock market do tomorrow?'' or ``will the next raven we see be
black?'' are already in the form of prediction. In these cases
all that needs to be done is to explicitly provide any relevant
historic data, such as stock market records or past weather
patterns, as a chronological input sequence and then look for
the natural continuations of these sequences. It should however
be noted that a simple formulation does not imply a simple
solution. For example, the chaotic nature of stock markets and
weather patterns make it extremely difficult to find the
correct continuation of this sequence, particularly more than a
few time steps ahead.

More formally, in the field of machine learning, sequence
prediction is concerned with finding the continuation
$x_{n+1}\in\X$ of any given sequence $x_1,...,x_n$. This may be
used to represent a wide range of abstract problems beyond the
obvious application to time series data such as historical
weather or stock patterns. For instance, (online) concept
learning, classification and regression can be regarded as
special cases of sequence prediction.

Concept learning involves categorizing objects into groups
which either do or don't possess a given property. More
formally, given a concept $C\subset\X$, it requires learning a
function $f_C$ such that for all $x$:
\[
f_C(x)=
  \begin{cases}
    1 & \mbox{if }x\in C\\
    0 & \mbox{if }x\notin C
  \end{cases}
\]
Solomonoff induction only deals with the problem of sequence
prediction; however, as we discuss in the next paragraph,
sequence prediction is general enough to also capture the
problem of concept learning, which itself is a specific case of
classification. Although the setup and interpretation of
classification using Solomonoff may be less intuitive than
using more traditional setups, the excellent performance and
generality of Solomonoff implies that theoretically it is
unnecessary to consider this problem separately.

In machine learning, classification is the problem of assigning
some given item $x$ to its correct class based on its
characteristics and previously seen training examples. In
classification we have data in the form of tuples containing a
point and its associated class $(x_{i},c_{i})$. The goal is to
correctly classify some new item $x_{n+1}$ by finding
$c_{n+1}$. As before, all data is provided sequentially with
the new point $x_{n+1}$ appended at the end. In other words,
the classification of $x$ becomes ``what is the next number in
the sequence $x_1c_1x_2c_2...x_{n}c_{n}x_{n+1}$?''. Technically
this could be regarded as a specific case of regression with
discrete function range, where the function we are estimating
maps the items to their respective classes.

Regression is the problem of finding the function that is
responsible for generating some given data points, often
accounting for some noise or imprecision. The data is a set of
(feature,value) tuples $\{(x_1,f(x_1))$,
$(x_2,f(x_2))$,....$(x_n,f(x_n))\}$. In machine learning this
problem is often tackled by constructing a function that is
the `best' estimate of the true function according to the data
seen so far. Alternatively, it can be formalized directly in
terms of sequential prediction by writing the input data as a
sequence and appending it with a new point $x_{n+1}$ for which
we want to find the functional value. In other words the
problem becomes: ``What is the next value in the sequence
$x_1,f(x_1),x_2,f(x_2),...x_n,f(x_n),x_{n+1},?$''. Although
this approach does not produce the function explicitly, it is
essentially equivalent, since $f(x)$ for any $x$ can be
obtained by choosing $x_{n+1}=x$.

\subsection{Prediction with Expert Advice versus Bayesian Learning} 

Prediction with expert advice is a modern approach to
prediction. In this setting it is assumed that there is some
large, possibly infinite, class of `experts' which make
predictions about the given data. The aim is to observe how
each of these experts perform and develop independent
predictions based on this performance. This is a general idea
and may be carried out in various ways. Perhaps the simplest
approach, known as follow the leader, is to keep track of which
expert has performed the best in the past and use its
prediction. If a new expert takes the lead, then
your predictions will switch to this new leading expert.
Naively the performance of an expert can be measured by simply
counting the number of errors in its predictions but in many
situations it is appropriate to use a loss function that
weighs some errors as worse than others.
A variant of this simple `follow the leader' concept is known
as `follow the perturbed leader' in which our predictions mimic
the leader most of the time but may switch to another with some
specified probability \cite{Hutter:05expertx}. This technique
gives a probability distribution rather than a deterministic
predictor which can be advantageous in many contexts.

The traditional Bayesian framework discussed in this article
uses a mixture model over a hypothesis or environment or model
class, which resembles the `follow the perturbed leader'
technique. This mixture model reflects our rational beliefs
about the continuation of a sequence given the performance of
each ``expert'' and, as we will see, performs very well
theoretically. Solomonoff induction uses the Bayesian framework
with the infinite class of ``experts'' given by all computable
environments. This means that there is always an expert that
performs well in any given environment which allows for good
performance without any problem-specific assumptions.

\subsection{No Free Lunch versus Occam's Razor}

This is in some way a contradiction to the well-known {\em
no-free-lunch theorems} which state that, when averaged over
all possible data sets, all learning algorithms perform equally
well, and actually, equally poorly \cite{Wolpert:97}. There
are several variations of the no-free-lunch theorem for
particular contexts but they all rely on the assumption that
for a general learner there is no underlying bias to exploit
because any observations are equally possible at any point. In
other words, any arbitrarily complex environments are just as
likely as simple ones, or entirely random data sets are just as
likely as structured data. This assumption is misguided and
seems absurd when applied to any real world situations. If
every raven we have ever seen has been black, does it really
seem equally plausible that there is equal chance that the next
raven we see will be black, or white, or half black half white,
or red etc. In life it is a necessity to make general
assumptions about the world and our observation sequences and
these assumptions generally perform well in practice.

To overcome the damning conclusion of these {\em no-free-lunch
theorems} in the context of concept learning, Mitchell
introduced the following inductive learning assumption which
formalizes our intuition and is essentially an implicit part of
our reasoning \cite{Mitchell:90}.

{\em ``Any hypothesis found to approximate the target function
well over a sufficiently large set of training examples will
also approximate the target function well over other unobserved
examples.''}

Similar assumptions can be made for other contexts but this
approach has been criticized as it essentially results in a
circular line of reasoning. Essentially we assume that
inductive reasoning works because it has worked in the past but
this reasoning is itself inductive and hence circular. Hume's
argument that this circularity invalidates inductive reasoning
is discussed further in Subsection~\ref{sec:Hume}. In
fact this inductive learning assumption is closely related to
what Hume called the principle of uniformity of nature. A
principle he said we implicitly, but invalidly, assume.

If we prescribe Occam's razor principle \cite{Ockham:90} to
select the simplest theory consistent with the training
examples and assume some general bias towards structured
environments, one can prove that inductive learning ``works''.
These assumptions are an integral part of our scientific
method. Whether they admit it or not, every scientist, and in
fact every person, is continuously using this implicit bias
towards simplicity and structure to some degree.

Any agent, animal or machine, must make use of underlying
structure in some form in order to learn. Although induction
inherently involves dealing with an uncertain environment for
which no hard guarantees can be made, it is clear that our
world is massively structured and that exploiting structure in
general is the best technique for performing well. By denying
the relevance of this structure {\em no-free-lunch theorems}
imply that general learning, and the concept of general
intelligence, is essentially futile, which contradicts our
experience. Solomonoff induction is witness to the possibility
of general learning, assuming only {\em some} structure in the
environment without having to specify which type of structure,
and using Occam's razor.

\subsection{Non-Monotonic Reasoning}

Non-monotonic reasoning is another concept that has been
discussed recently in relation to induction. This concept
attempts to solve the problem of formalizing common sense
logic. When artificial intelligence researchers attempted to
capture everyday statements of inference using classical logic
they began to realize this was a difficult if not impossible
task. The problem arises largely because of the implicit
assumption of normality we often make to exclude exceptional
circumstances. For example, it would be perfectly acceptable to
make a statement such as ``the car starts when you turn the key
in the ignition'' \cite{Gabbay:11}. Therefore if we let $F(x)$
be the predicate that we turn the key in the ignition in car
$x,$ and $S(x)$ be the predicate that $x$ starts, then the
previous sentence would be represented by the logical statement
$F(x)\limpl S(x)$. Of course there are many reasons why this
might not be correct such as the fact that the car $x$ has no
fuel or a mechanical problem. But these exceptions do not stop
us making these types of statements because it is implicity
assumed that this statement may only hold under normal
conditions.

This assumption of normal conditions also leads to a logic
that is non-monotonic in its arguments. Normally if the
statement $A\limpl C$ holds, then it follows logically that
$A\wedge B\limpl C$. However this rule may no longer hold
using `normal case' reasoning. If $G(x)$ is the predicate that
$x$ has no fuel then, although the statement $F(x)\limpl
S(x)$ is (normally) true, $F(x)\wedge G(x)\limpl S(x)$ is
(normally) not true, since the car will not start without fuel.
Another example is that a general rule in our knowledge base
may be that ``birds can fly''. Therefore if $x$ is a bird it is
natural to assume that $x$ can fly; however if $x$ is a bird
and $x$ is a penguin then it is no longer correct to say that
$x$ can fly.

\subsection{Solomonoff Induction}

Solomonoff induction \cite{Solomonoff:64} bypasses this issue
entirely by avoiding the use of strict logical syntax which
seems to be an inadequate tool for dealing with any reasonably
complex or real-world environment. Non-monotonic statements
such as the examples shown can be programmed in a variety of
ways to effectively deal with `the normal case' and an
arbitrary number of exceptional cases. This means that there
exists a computable environment in Solomonoff's universal class
which will effectively describe the problem. The
non-monotonicity of the environment will certainly affect its
complexity and hence its prior but a simple non-monotonic
environment will still have a reasonably high prior since there
will be a reasonably short way of expressing it. More generally
the complexity, and hence prior (see
Subsection~\ref{sec:uprior}), of a non-monotonic environment
will depend on the variety and number of exceptions to the
general rules, but this seems to be a desirable property to
have. The implicit assumption of normality we use is due to our
prior experience and knowledge of the real world. Initially,
for an agent acting in an unknown environment, it seems
reasonable that upon being told a general rule, it should
assume the rule to hold in all cases and then learn the
exceptional cases as they are observed or inferred. This is
essentially how Solomonoff induction behaves.

Because of the fundamental way in which Solomonoff's universal
induction scheme continuously learns and improves from its
experience, it may be argued that any useful computable
approach to induction in one way or another approximates
Solomonoff induction. In any case it appears to compare
well with the above approaches. The major issue remains its
incomputability and the difficulty of approximating Solomonoff
in reasonably complex environments. This is discussed further
in Section~\ref{cha:app}.

\section{Probability}\label{cha:prob}

In order to fully appreciate the Bayesian framework it is
important to have some understanding of the theory of
probability that it is based on. Probability theory has had a
long and contentious history
\cite{Good:83,Jaynes:03,McGrayne:11}. Even today probability
theory divides the scientific community with several competing
schools of thought which stem largely from the different
methods of dealing with uncertainty as it appears in different
areas of science. The most popular of these are objective,
subjective and frequentist which reflect fundamentally
different philosophical interpretations of what probability
means. Surprisingly it turns out that these interpretations
lead to the same set of axioms and therefore these
philosophical differences are of little importance in practical
applications. It is however worth considering these differences
when looking at our motivation in the context of induction.

In the following $\Omega$ is used to denote the sample space
which is the set of all possible outcomes. An event
$E\subset\Omega$ is said to occur if the outcome is in $E$. For
instance when throwing a die the sample space $\Omega$ is
$\{1,2,3,4,5,6\}$ and an event $E$ is some specific subset of
these outcomes. For instance, the even numbers is the event
$E=\{2,4,6\}$ and the set of numbers less than $3$ is the event
$E=\{1,2\}$.

\subsection{Frequentist}\label{sec:freq}

The frequentist approach is the most intuitive interpretation
of probabilities, however it has several crippling drawbacks and
is not applicable in many situations where we would like it to
be. A frequentist defines the probability of an event as the
limiting frequency of this event relative to the entire sample
space $\Omega$. Formally if $k(n)$ is the number of times that
event $E$ occurs in $n$ trials then
$P(E):=\lim_{n\to\infty}[k(n)/n]$. For example when throwing a
die the probability of throwing a $6$ is defined as the ratio of
the number of throws that come up $6$ to the number of throws
in total as the number of throws goes to infinity. After many
throws we expect this number to be close to $\frs{1}{6}$. This
is often one of the ways the concept of probability is taught,
which is part of the reason that it appeals to our intuition.
However when examined more closely it becomes apparent that
this definition is problematic.

No matter how large $n$ gets there is no guarantee that
$k(n)/n$ will converge to $P(E)$. Even if the die is thrown a
million times it is conceivable although extremely unlikely
that every roll will produce a $6$ or that half the rolls will
produce a $6$. The best we can say is that as $n$ increases,
the probability that $k(n)/n$ is arbitrarily close to $P(E)$
also increases and will eventually get arbitrarily close to
$1$. Formally this is stated as $k(n)/n$ converges to $P(E)$
with probability $1$. Unfortunately this gives rise to a
circularity as the concept of probability is then used in
defining probability.

Another problem with the frequentist approach is that there are
many situations where it is not applicable. Consider the
betting odds in a horse race. If the odds on a horse are, for
instance, $3:1$ this is equivalent to saying that the
probability that the horse will win is $0.25$. This is
certainly not the same as saying that the horse has won $1$ in
every $4$ previous races. Instead it represents the bookies
belief that the horse will win which depends on many factors.
This probability as belief interpretation is the basis of the
subjectivist's understanding.

The error here may appear to be in associating probabilities
with betting odds and it could be argued that strictly speaking
the probability of the horse winning should be defined as the
ratio of wins to overall races in the past, but this idea
quickly becomes inconsistent. Clearly it makes no sense to
equally weigh every race the horse has been in to find the
probability of the horse winning this particular race. The
races might therefore be restricted to those held on the same
track and against the same horses, but since the weather and age
of the horse might also be a factor there would be no other
races with which to compare. This choice of reference class
poses a very real problem in some practical situations such as
medical diagnosis \cite{Reichenbach:49}. The frequency approach
is only really applicable in situations where we can draw a
large number of samples from a distribution that is
independent and identically distributed (i.i.d.) such as flipping a
coin.

\subsection{Objectivist}\label{sec:obj}

The objectivist interpretation is that probabilities are real
properties of objects and of the world and therefore the
objectivist believes that the world actually involves
inherently random processes. This point of view has been
largely supported by the success of quantum physics which
states that there is true randomness present at a sub-atomic
level. The most widely accepted set of axioms for objective
probabilities are due to Kolmogorov \cite{Kolmogorov:33} and
are given here.

\paradot{Kolmogorov's Probability axioms}
\begin{itemize}\parskip=0ex\parsep=0ex\itemsep=0ex
\item If $A$ and $B$ are events, then the intersection $A\cap B$, the
union $A\cup B$, and the difference $A\setminus B$ are also events.
\item The sample space $\Omega$ and the empty set $\{\}$ are events.
\item There is a function $P$ that assigns non negative real numbers, called
probabilities, to each event.
\item $P(\Omega)=1$ and $P(\{\})=0$.
\item $P(A\cup B)=P(A)+P(B)-P(A\cap B)$
\item For a decreasing sequence $A_1\supset A_2\supset A_3...$of
events with $\cap_{n}A_{n}=\{\}$ we have $\lim_{n\to\infty}P(A_n)=0$
\end{itemize}
In addition to the axioms there is the important definition of
conditional probability. If $A$ and $B$ are events with
$P(A)>0$ then the probability that event $B$ will occur under
the condition that event $A$ occurred is
\[
  P(B|A) \;=\; \frac{P(A\cap B)}{P(A)}
\]
One of the problems with these axioms is that they only
uniquely determine values for the null event and the event of
the entire probability space. Although there are general
principles for assigning values to other events, finding a
universal formal method has been problematic. Applying general
principles often requires some degree of subjectivity which can
lead to debate. Kolmogorov complexity which is examined later
provides a promising universal framework.

It has been argued \cite{Hutter:04uaibook,Schmidhuber:97brauer}
that the objective interpretation places too much faith in the
ultimate truth of the quantum physics model. A simple example
of randomness being incorrectly attributed to a process is the
flipping of a coin. This is the standard analogy used in almost
any situation where two outcomes occur with probability $0.5$
each, with heads and tails representing the respective
outcomes. This is because when we flip a coin the probability
of heads is, for all practical purposes, $0.5$. Even in this
article we used this example to represent a truly stochastic
process, but in reality the probability of heads is actually
(close to) $1$ or $0$ the moment the coin leaves your finger.
This is because the process is not inherently random and if the
exact initial conditions are known then the outcome can be
calculated by applying the laws of physics. This statement is
somewhat questionable as we may debate that an unknown breeze
may affect the outcome or that our calculations would also need
to consider the exact point that the coin lands which, if it
was caught, would depend on the person. These objections are
mute if we consider exact knowledge of initial conditions to
include all local weather conditions and the persons state of
mind. Without going into the question of free will the point is
simply that we often use randomness to account for uncertainty
in a model or a lack of adequate knowledge. Perhaps quantum
physics is analogous to this in that although the model is
currently very successful, there may be a time when the
`inherent' randomness can be deterministically accounted for by
a more accurate model.

In cases where the data is i.i.d., the objective probability is
still identified with the limiting frequency, which is why
these interpretations coincide for these cases. It is also
possible to derive these axioms from the limiting frequency
definition, however by using these axioms as a starting point
the issues encountered by the frequentist are avoided.

\subsection{Subjectivist}\label{sec:subj}

It is the subjectivist interpretation of probability that is
most relevant in the context of induction, particularity in
relation to agent based learning. The subjectivist interprets a
probability of an event as a degree of belief in the event
occurring and when any agent, human included, is attempting to
learn about its environment and act optimally it is exactly
this degree of belief that is important.

If a probability defines a degree of belief it must be
subjective and therefore may differ from agent to agent. This
may seem unscientific or unsatisfactory but when examined
philosophically this interpretation has a strong case. To see
this, consider Scott and Fred gambling in Las Vegas. While
playing roulette they observe that the ball lands on $0$ an
abnormally high number of times causing them to lose
significantly. Given that they are in a large well-known
Casino, Fred thinks nothing of this abnormality believing that
they have probably just been very unlucky. Scott on the other
hand knows some of the employees at this particular Casino and
has heard rumors of corruption on the roulette tables. This
extra information can be thought of as observations that when
combined with the statistical abnormality of the roulette table
raises Scott's belief that they have been victim to foul play.
It is inevitable that, consciously or not, our analysis and
interpretation of any situation will be biased by our own
beliefs, experience and knowledge.

In a very formal sense this means that our probabilities are a
function of our entire previous personal history and this is
exactly how Solomonoff's prediction scheme can be used. As a
simple example consider Fred and Scott each independently
drawing balls from an urn, with replacement, which contains
black and white balls in an unknown ratio. Imagine that Fred
draws $50$ white balls and $20$ black balls while Scott draws
$30$ white balls and $40$ black balls. This is possible for any
true ratio as long as there is a positive fraction of both
black and white. Clearly Fred will believe that the ratio of
white to black is approximately $5:2$ while Scott will believe
that it is approximately $3:4$. The point is that both of these
beliefs are completely valid given their respective
observations regardless of the true ratio.

Although we may accept that probabilities are subjective it is
vital that there is a formal system that specifies how to
update and manipulate these belief values. It is here that the
subjective interpretation of probability has faced many
criticisms as it was argued that subjective belief values don't
obey formal mathematical rules or that the rules they do obey
are also subjective making formalization difficult or
impossible. It is a surprising and major result that any
rational agent must update its beliefs by a unique system which
coincides with that of limiting frequencies and objective
probabilities.

The most intuitive justification for this is from a Dutch book
argument which shows that if an agent's beliefs are
inconsistent (contradict the axioms) then a set of bets can be
formulated which the agent finds favorable according to its
beliefs but which guarantees that it will lose. The Dutch book
argument is however not rigorous and there are several
objections to it \cite{Earman:93}. The main issue rests on the
implicit assumption that belief states uniquely define betting
behavior which has been called into question since there are
other psychological factors which can have an affect. For
example in a game of poker it is often rational for a player to
bet an amount that does not reflect his belief in winning the
hand precisely because he is trying to bluff or convey a
weak hand \cite{Schick:86}. In 1946 Cox published a theorem
that gave a formal rigorous justification that {\em ``if
degrees of plausibility are represented by real numbers, then
there is a uniquely determined set of quantitative rules for
conducting inference''} \cite{Jaynes:03} and that this set of
rules is the same as those given by the standard probability
axioms.

\paradot{Cox's axioms for beliefs}
\begin{itemize}\parskip=0ex\parsep=0ex\itemsep=0ex
\item The degree of belief in an event $B$, given that
    event $A$ has occurred can be characterized by a
    real-valued function $Bel(B|A)$.
\item $Bel(\Omega\setminus B|A)$ is a twice differentiable
    function of $Bel(B|A)$ for $A\neq\{\}$.
\item $Bel(B\cap C|A)$ is a twice differentiable function
    of $Bel(C|B\cap A)$ and $Bel(B|A)$ for $B\cap
    A\neq\{\}$.
\end{itemize}
This unification and verification of the probability axioms is
a significant result which allows us to view the frequentist
definition as a special case of the subjectivist
interpretation. This means that the intuitively satisfying
aspect of the frequentist interpretation is not lost but now
obtains a new flavor. Consider again the case of determining
the ratio of black to white balls in an urn through repeated
sampling with replacement where the true ratio is $1:1$. As the
urn is repeatedly sampled the relative frequency and hence
subjective belief that the next ball is white will converge
with probability $1$ to $0.5$. Although this is the correct
probability it is important to realize that it is still a
belief and not an inherent property. In the unlikely but
possible event that a white ball is sampled $1000$ times the
subjective probability/belief that the next ball will be white
would be very close to $1$.

This understanding of probability can be troubling as it
suggests that we can never be certain of any truth about
reality, however this corresponds exactly with the philosophy
of science. In science it is not possible to ever prove a
hypothesis, it is only possible to disprove it. No matter how
much evidence there is for a hypothesis it will never be enough
to make its truth certain. What are often stated as physical
laws are actually only strongly believed and heavily tested
hypotheses. Science is not impeded by this fact however. On the
contrary, it allows for constant questioning and progress in
the field and, although models may never be completely proven,
it does not stop them being usefully applied.

\section{Bayesianism for Prediction}\label{cha:bayes}

To fully appreciate the historical attempts to solve the
problem of induction and the corresponding discussions which
fueled the field it is necessary to first understand the
Bayesian framework. But before explaining the mechanics of the
Bayesian framework it is worth having a brief look at what it
means to be a Bayesian. Giving a precise explanation is
difficult due to the various interpretations of Bayesianism
\cite{Good:71,Good:83}, however all Bayesians share some core
concepts.

Being a Bayesian is often simply associated with using Bayes
formula but this is a gross simplification. Although Bayes
formula plays an important role in the Bayesian framework it is
not unique to Bayesians. The rule is directly derived from the
axioms of probability and therefore its correctness is no more
debatable than that of the axioms of probability.

More important to Bayesianism is Cox's result that a rational
belief system must obey the standard probability axioms. This
is because a Bayesian is a subjectivist, believing that our
beliefs and hence probabilities are a result of our personal
history. In other works, what we believe today depends on what
we believed yesterday and anything we have learnt since
yesterday. What we believed yesterday depends on what we
believed the day before, and so forth. Two individuals with
very different histories may therefore hold different beliefs
about the same event. This means that the probability of the
event for each individual can be validly different from each
other as long as they both updated their beliefs in a
rationally consistent manner.

This rational updating process is at the core of Bayesianism.
It may seem unscientific that different individuals can assign
distinct yet valid probabilities but, as we have seen in
Subsection \ref{sec:subj}, this can be quite reasonable. There
is a strong link to the frequentist approach here. If the two
individuals are given the same observations, or at least
observations from the same source, then their beliefs should
eventually converge because their frequentist estimate should
converge. Philosophically speaking, however, different
individuals will never observe precisely the same observations.
Each human has a unique experience of the world around them and
therefore their beliefs will never be identical.

A Bayesian's belief about events is governed by beliefs in the
possible causes of those events. Everything we see has many
possible explanations although we may only consider a few of
them to be plausible.\textbf{ }To be able to update beliefs
consistently a Bayesian must first decide on the set of all
explanations that may be possible. When considering a specific
experiment this set of explanations, or hypotheses, need only
explain the observations pertaining to the experiment. For
example when flipping a coin to find its bias, the hypotheses
may simply be all possible biases of the coin. For universal
induction, we are interested in finding the true governing
process behind our entire reality and to do this we consider
all possible worlds in a certain sense. No matter what the
problem is we can always consider it to consist of an agent in
some unknown environment. In the coin example all irrelevant
information is discarded and the environment simply consists of
observing coin flips. It is useful to keep this general setup
in mind throughout this article.

Lastly, the agent must have some prior belief in these
explanations before the updating process begins. In other words
before any observations have been made. Our beliefs today
depend on beliefs yesterday which depend on the day before. But
at some point there is no `day before' which is why some
initial belief is required to begin the process. Over a long
enough period these initial beliefs will be `washed out' but
realistically they are important and should be chosen sensibly.

Summing up, a Bayesian holds beliefs about any possible cause
of an event. These beliefs depend on all previously obtained
information and are therefore subjective. A belief system that
is entirely consistent with the Bayesian framework is obviously
unrealistic as a model for human reasoning as this would
require perfect logical updating at every instance as we
continuously receive new information. There are also emotional
and psychological factors that come into play for humans.
Rather this is an idealized goal, or gold standard, which a
Bayesian thinks we should strive for if we are to be completely
rational.

\subsection{Notation}

In order to examine some of the more technical content and
results it is necessary to first establish some notation.
Throughout the article $\X$ will represent the alphabet of the
observation space being considered. This is simply the set of
characters used to encode the observations in a particular
situation. For example when flipping a coin and observing heads
or tails, $\X=\{Head,\, Tail\}$ or $\{h,\, t\}$ or $\{1,0\}$.
An observation sequence is encoded as a string over the
alphabet which is usually denoted $x$. In some cases we are
interested in the length of $x$ or some subsection of $x$.
$x_{1:n}$ denotes a string of length $n$ or, depending on
context, the first $n$ bits of $x$. $x_m$ denotes the $m$th bit
of $x$. $x_{<n}$ is the same as $x_{1:n-1}$, meaning all bits
before the $n$th bit. $\X^*$ denotes the set of strings in this
alphabet that have finite length. Therefore $x\in\X^*$ means
$x$ is any possible finite observation sequence.

In the Bayesian framework we deal with various environments,
the class of all possible considered environments is denoted
$\M$. It is assumed that one of these environments is the true
environment which is denoted $H_\mu$ or simply $\mu$. These
hypotheses each specify distributions over strings $x$ in the
corresponding alphabet. This means that
$\nu(x):=H_\nu(x):=P(x|H_\nu)$ can be thought of as the
probability of $x$ according to environment $H_\nu$. The
concept of conditional probability is the same for these
distributions with $H_\nu(x|y)$ being the probability of
observing $x$ given that $y$ was observed, under hypothesis
$H_\nu$. We have an initial belief in each of the environments
$H_\nu$ in $\M$ which is denoted $w_\nu=P(H_\nu)$. Note that
probability and belief are used interchangeably in the Bayesian
framework due to its subjectivist perspective. We may also be
interested in our posterior belief in an environment $H_\nu$
after observing $x$. This is denoted $w_\nu(x):=P(H_\nu|x)$.

\subsection{Thomas Bayes}

Reverend Thomas Bayes is a highly enigmatic figure in the
history of mathematics. Very little is known of his life and
few of his manuscripts survived. He only published one
mathematical paper in his life, yet was elected to be a fellow
of the royal society of London. On the 17th April 1761 Bayes
died leaving unpublished an essay that would have a profound
impact on the minds and methodologies of countless future
scientists. This was not a discovery about the nature of our
universe but rather a framework for how we can draw inferences
about the natural world, essentially about how to rationally
approach science. It wasn't until two years later that Bayes'
friend Richard Price posthumously published his ``Essay towards
solving a problem in the doctrine of chances''
\cite{Bayes:1763} and it wasn't until many decades later that
it began to have any major influence.

The problem as Bayes explicitly states it at the beginning of
his essay is this:

{\em Given:} The number of times in which an unknown event has
happened and failed

{\em Required:} The chance of its happening in a single trial
lies somewhere between any two degrees of probability that can
be named.

Intuitively Bayes was looking at the problem of inverse
inference. At the time almost all applications of probability
were to do with direct inference \cite{Zabell:89}. These were
cases where the initial mechanism and probabilities were known
and the probability of some outcome was to be deduced. Bayes
was instead interested in the case where experimental evidence
was known but the true nature of the environment was unknown.
In particular, given the number of successes and failures of an
experiment what is the probability of success on the next
trial?

The history of the development of inductive reasoning stretches
over centuries with contributions from many great minds. It was
actually through the apparently independent work of Laplace a
decade later that the techniques of inverse probability gained
widespread acceptance. Later we look at some of this history of
inductive reasoning but here the focus is on the mechanics of
the Bayesian framework which underlies most of the work in this
field.

\subsection{Models, Hypotheses and Environments}

Whether we are testing hypotheses or modeling the world around
us, science is fundamentally about understanding our
environment in order to benefit from it in the future.
Admittedly this statement may not be universally accepted, but
it is from this utilitarian perspective that this article is
written. This perspective is certainly influenced by the agent
based setting of artificial intelligence.
The entire field of economics, for example, is based on this
assumption. In any case this abstract view of science is useful
in order to see the relation with inductive inference. We can
regard ourselves as agents maximizing some unknown reward
function or utility. It is the complete generality of this
utility that makes this assumption realistic. We may not be
simply maximizing our bank balance, but the very fact that we
have preferences between outcomes implies some inbuilt,
possibly complex, reward function. Under this perspective,
science is a tool for prediction which allows us to make
decisions that maximize our utility; and models, environments
and hypothesis are tools for science.

In looking at Bayesianism, it is important to realize that
hypotheses, models and environments all express essentially the
same concept. Although they may seem intuitively different they
each express an explanation of some phenomenon at various
levels of abstraction. A hypothesis is generally in relation to
a specific experiment and is therefore quite a local
explanation. A model is often used to explain a larger set of
observations such as the movement of celestial bodies or the
flow of traffic in a city, although models may be as
comprehensive or specific as we like. Our environment may be
thought of as a comprehensive model of our world, yet we can
also use the term in reference to specific information relevant
for some context such as a particular game environment. For
example a chess environment consists of the current
configuration of the pieces on the board and the rules that
govern the game. The point here is that for the purposes of
Bayesian learning no distinction is made between models,
environments and hypotheses.

When thinking about a model or environment it is common to
think of some meaningful underlying mechanism or program that
governs the output we receive in the form of observations,
however this is not necessary. At an abstract level, any
complete hypothesis can be thought of as specifying a
probability distribution over some observation set. In general
the observations may not be independent so this distribution
must be with respect to any previous observations generated by
the environment. We consider this data to be encoded in a
string over some finite alphabet which we denote $\X$, and
the distribution is then over this alphabet. If the environment
is deterministic then this distribution is simply concentrated
to a single character of the alphabet at each step resulting in
the probability of the correct string being one, or certain,
and the probability of any other string being zero.

To illustrate this consider the simple environment of a fair
coin being flipped. There are only two possible observations,
either heads or tails so the alphabet $\X$ is $\{h,t\}$. In
this case the observations are independent, so regardless of
the observations seen so far the distribution given by this
environment remains the same. For notation let $E$ be some
history sequence over $\{h,t\}$ and let $H_C$ be the
distribution. Then we have, for any history $E$, $H_C(h|E)=0.5$
and $H_C(t|E)=0.5$ . If we consider the environment to instead
be an ordered deck of cards $H_D$ then it is clear that if
$Z$ is some initial portion of the deck and $a$ is the card
following $Z$ then $H_D(a|Z)=1$ and $H_D(\bar a|Z)=0$
where $\overline a$ is any other card in the deck. Since the
card $a$ following $Z$ is obviously dependent on $Z$, it is
clear that this distribution changes with the history $Z$. We
call data sets that are drawn from independent distributions
such as the coin flip, independent and identically distributed
data, or i.i.d.\ for short.

In order to operate the Bayesian framework it is necessary to
assume there is some class of hypotheses $\M$ under
consideration that contains the correct hypothesis $H_\mu$. We
will see that a countable class is sufficient for universal
induction. Therefore, unless otherwise indicated, we (can)
assume that $\M$ is countable. Each of these hypotheses must be
assigned some prior which represents the belief in this
hypothesis before any data has been observed. For each
$H_\nu\in\M$ we denote this prior by $w_\nu=P(H_\nu)$. This is
something we do to a certain extent naturally. Imagine you see
something unusual on the way to work such as an apartment block
surrounded by police cars and an ambulance. Naturally you might
consider a number of different explanations for this. Perhaps
there was a break in, or a murder, or a drug bust. There is
essentially a countless number of possible explanations you
might consider which is analogous to the model class.
Presumably one of these is correct and you would not consider
them all to be equally likely. Your plausibility of each
corresponds to the priors $w_\nu$.

In order to be rigorous it is required that the hypothesis
class be mutually exclusive and that the priors sum to one.
That is $\sum_{\nu\in\M}w_\nu=1$. These requirements are simply
equivalent to requiring that our belief system is rationally
consistent. The prior belief in a particular observation string
is the weighted sum of the beliefs in the string given by each
hypothesis with weight proportional to the belief in that
hypothesis. Formally
$P(x_{1:n})=\sum_{H_\nu\in\M}P(x_{1:n}|H_\nu)P(H_\nu)$ where
$P(H_\nu)=w_\nu$. This is a key concept, known as Bayes
mixture, or marginal likelihood, which will be explained
further. Note that the implicit dependence on ``background
knowledge'' $\M$.

\subsection{Bayes Theorem}
Bayes theorem is used to update the belief in a hypothesis
according to the observed data. This formula is easily derived
from the definition of conditional probability.
\[
  P(A|B) \;=\; \frac{P(B|A)P(A)}{P(B)}
\]
A set of events is mutually exclusive and complete when any
event in the sample space $\Omega$ must belong to one and only
one event from this set. For example, any event $E$ along with
it complement $\overline{E}=\Omega\setminus E$ form a mutually
exclusive and complete set. Imagine you're throwing a die and
$E$ is the event of a $6$. Obviously any throw must be in
either the set $\{6\}$ or its complement $\{1,2,3,4,5\}$. If
$\Lambda$ is a mutually exclusive and complete countable set of
events, then $P(B)=\sum_{C\in\Lambda}P(B|C)P(C)$. This can be
derived by using all and only Kolmogorov's axioms of
probability and the definition of conditional probability
stated in Section \ref{sec:subj}. Therefore Bayes theorem can
be given in the form
\[
  P(A|B) \;=\; \frac{P(B|A)P(A)}{\sum_{C\in\Lambda}P(B|C)P(C)}
\]
Now let $H_\nu$ be a hypothesis from class $\M$, and $x_{1:n}$
is be observational data. Since the model class $\M$ is
required to be mutually exclusive and complete, Bayes formula
can be expressed as follows
\[
  P(H_\nu|x_{1:n}) \;=\; \frac{P(x_{1:n}|H_\nu)P(H_\nu)}{\sum_{H_{i}\in\M}P(x_{1:n}|H_{i})P(H_{i})}
\]
This is the posterior belief in $H_\nu$ conditioned on data
$x_{1:n}$ is denoted $w_\nu(x_{1:n})$. The term $P(H_\nu)$ is
the prior belief $w_\nu$ and the term $P(x_{1:n}|H_\nu)$ is
known as the likelihood. This likelihood is the probability of
seeing the data $x_{1:n}$ if hypothesis $H_\nu$ is the true
environment. But this is exactly how we defined the
distribution given by hypothesis $H_\nu$ and therefore
$P(x_{1:n}|H_\nu)=H_\nu(x_{1:n})$.

\subsection{Partial Hypotheses}

So far we have only considered complete hypotheses. These
specify complete environments and therefore uniquely determine
a probability distribution. In many cases however, we have
hypotheses that only partially specify the environment.
Instead they represent the set of all environments that
satisfy some property, such as the property that some given
statement is true.

For example consider a coin with some unknown bias $\t$,
representing the probability the coin will land heads. We now
flip the coin repeatedly to find its exact bias. An example of
a complete hypothesis is the statement ``the coin will land on
heads $70\%$ of the time'', or $\t=0.7$. An example of an
incomplete hypothesis is the statement ``the coin will land
heads somewhere between $50\%$ of the time and $75\%$ of the
time'', or $\t\in(0.5,0.75)$. It's clear that the first
hypothesis is a specific case of the second hypothesis. Now
consider the statement ``all ravens are black''. This is a
partial hypothesis that consists of any environment where all
ravens are black, or where there are no non-black ravens. This
will be relevant when we examine the black ravens paradox.

It is important to be careful when dealing with these partial
hypotheses. We showed that for a complete hypothesis $H_\nu$ we
have $P(x_{1:n}|H_\nu)=H_\nu(x_{1:n})$. For a partial
hypothesis $H_p$ this no longer holds since $H_p$ refers to a
set of distributions and hence $H_p(x_{1:n})$ is undefined.

\subsection{Sequence Prediction}

From a metaphysical perspective it may be argued that our
ultimate goal is to understand the true nature of the universe,
or in this context to know which of the environment
distributions is the true distribution $H_\mu$, which will
simply be denoted $\mu$ from now on. But since we are taking an
agent based point of view, our aim is to make optimal decisions
in terms of maximizing some utility function. From this more
pragmatic perspective the primary concern is to make accurate
predictions about the environment. This is called a prequential
or transductive setting. Since the optimal predictions are
given by the true distribution these approaches are not
radically different; however, as we will see, it is not
necessary to perform the intermediate step of identifying the
correct distribution in order to make good predictions.

In some sense it is predictive power that has always been the
primary function of science and reasoning. We know that
realistically our scientific models may never be completely
correct, however they are considered successful when they yield
predictions that are accurate enough for our current purposes.

The goal is to make predictions that are optimal given the
information available. Ideally this information extends to
knowledge of the true environment $\mu$ in which case the
optimal prediction is simply defined by this distribution,
however this is rarely the case. Instead predictions must be
based on some estimate of the true distribution which reflects
an educated guess $\rho$ of $\mu$. Let all previous
observational data be in the form of the string $x\in\X^*$. Obtaining the posterior or predictive $\rho$-probability
that the next observation will be $a\in\X$ is given by the
conditional probability under $\rho$. Formally,
$\rho(a|x):=\rho(xa)/\rho(x)$. In the Bayesian framework this
estimation is given by Bayes mixture $\xi$.

\subsection{Bayes Mixture}

If our class of hypotheses is countable then we can use the
weighted average of all our hypotheses by their respective
priors as our best guess estimation $\rho$ of the true
environment $\mu$. This can be thought of as the subjective
probability distribution as described above. Formally it is
called Bayes Mixture and defined as follows
\[
  \xi(x) \;:=\; \sum_{\nu\in\M}w_\nu\cdot \nu(x)
\]
This definition makes perfect intuitive sense. The contribution
of an environment $\nu$ to the prediction of an observation $x$
is a combination of the prediction of $x$ under this
environment $\nu(x)$ and the belief in this environment
$w_\nu$. Imagine some environment $q$ predicts $x$ with
certainty but your belief in this environment $w_{q}$ is small,
while all other environments in $\M$ predict $x$ with small
probability. We would expect, as Bayes mixture implies, that
the resulting probability of $x$ will remain small although
larger than if $q$ was not considered at all.

Continuing with the poker example above, your opponent's hand
may be thought of as the unknown environment. The opponent may
hold any hand and each of these corresponds to a particular
hypothesis $\nu$ which you belief with probability $w_\nu$. In
this case consider the observation $x$ to be the event that you
win the hand. If there are still cards to be dealt then the
prediction $\nu(x)$ may be uncertain depending on how likely
the remaining cards are to change the outcome. For example if you
have three aces and $\nu$ is the hypothesis that the opponent
has four diamonds, with one card to come, then
$\nu(x)\approx\frs{3}{4}$ since there is approximately
$\frs{1}{4}$ chance that the next card will be a diamond which
will make your opponents diamond flush beat your three aces. If
there are no remaining cards then $\nu(x)$ is simply one or
zero depending on whether your hand is better or worse than
hand $\nu$ respectively. Your beliefs $w_\nu$ may depend on
psychological factors and previous betting but your final
decision, if rational, should involve the estimation $\xi$ over
the various possible environments. Even if you have a sure
belief that an opponent has a particular hand $\mu$ then this
system remains consistent since $\xi$ simply becomes the
distribution given by $\mu$. This is because $w_\mu=1$,
$w_\nu=0$ for all $\nu\neq \mu$ and therefore $\xi(x)=\mu(x)$.

The probability of some observation under this Bayes mixture
estimation can be thought of as its subjective probability
since it depends on the priors which reflect our personal
belief in the hypotheses before any data has been observed. If
our belief $w_\nu=0$ for some $\nu$, it does not contribute to
$\xi$ and could equally well be dropped from $\M$. Therefore,
without loss of generality and consistent with Epicurus, we
assume $w_\nu>0$ from now on. An important mathematical
property of this mixture model is its dominance.
\[
  \xi(x)\ge w_\nu\cdot \nu(x)\quad\forall x\mbox{ and }\forall \nu\in\M,
  \qmbox{in particular}\xi(x)\ge w_\mu\cdot \mu(x)
\]
This means that the probability of a particular observation
under Bayes mixture is at least as great as its probability
under any particular hypothesis in proportion to the prior
belief in that hypothesis. This is trivial to see since the
probability under Bayes mix is simply obtained by summing the
probabilities under each hypothesis proportional to its prior
and these are all non-negative. In particular this result
applies to the true distribution $\mu$. This property is
crucial in proving the following convergence results.

\subsection{Expectation}

Since our predications deal with possibly stochastic
environments, expectation is an important concept in examining
performance. When there is random chance involved in what
rewards are gained, it is difficult to make guarantees about the
effect of a single action. The action that receives a higher
reward in this instance may not be optimal in the long run. As
long as we have arbitrarily many tries, the best strategy is to
choose the action that maximizes the expected reward. This is
particularly relevant in relation to agent based learning,

In general, expectation is defined for some function
$f:\,\X^n\to\mathbb{R}$, which assigns a real value to an
observation sequence of any length in the following way:
\[
  \E[f] \;=\; \sum_{x_{1:n}\in\X^n}\mu(x_{1:n})f(x_{1:n})
\]
This can be thought of as the average value of a function under
the true distribution. When we talk about maximizing an agent's
expected reward the function being considered is the agent's
utility function and this is generally the most important
value. For example, when given the choice between a certain
$\$10$ or a $50\%$ chance at $\$100$ the rational choice is to
take the latter option as it maximizes one's expectation,
assuming monetary value defines utility. Expectation is an
essential concept for making good decisions in any stochastic
environment.

In poker, for example, a good player uses expectation
continuously, although the calculations may eventually become
instinctual to a large degree. In general a player's decision
to continue in a hand depends on whether the expected return is
larger than the amount the player must invest in the hand. In
the case of Texas hold'em poker the true environment is the
distribution given by the shuffled deck and the function is the
expected return on some sequence of communal cards. It should
also be noted that this is not a stationary environment: the
distribution changes conditioned on the new information
available in the communal cards.

\subsection{Convergence Results}\label{sec:conv}

For the Bayesian mixture to be useful it is important that it
performs well. As the accuracy of predictions is the primary
concern, the performance of a distribution is measured by how
close its predictions are to those of the true environment
distribution. The analysis of this performance varies depending
on whether the true environment is deterministic or stochastic.

\paradot{Deterministic}
In the deterministic setting the accuracy is easier to
determine: As an observation either will or won't be observed,
there is no uncertainty. For a deterministic environment it is
sufficient to know the unique observation sequence $\alpha$
that must be generated, since it contains all the information
of the environment. Formally $\mu(\alpha_{1:n})=1$ for all $n$
where $\alpha_{1:n}$ is the initial $n$ elements of $\alpha$,
and $\mu(x)=0$ for any $x$ that is not a prefix of $\alpha$,
i.e.\ there is no $n$ such that $x=\alpha_{1:n}$. In this
deterministic case the following results hold
\[
  \sum_{t=1}^\infty|1-\xi(\alpha_t|\alpha_{<t})|\leq\ln(w_\alpha^{-1})\;<\;\infty
  \qmbox{and} \xi(\alpha_{t:n}|\alpha_t) \to 1 \mbox{ for } n\ge t\to\infty
\]
Although the true distribution is deterministic, and perhaps
even all environments in the hypothesis class $\M$, this does
not imply that Bayes mixture $\xi$ will be deterministic (see
Subsection~\ref{sec:det}).

Since an infinite sum of positive numbers can only be finite if
they tend to zero, this result shows that the probability that
the next observation will be predicted correctly under $\xi$,
given all previous observations, converges rapidly to one. The
sum of the probabilities that an incorrect, or off sequence,
observation is predicted is bounded $\ln(w_\alpha^{-1})$
depending on our prior belief in the true environment $\alpha$.
If our prior belief is one, the true environment is known, then
this constant is zero, meaning that we never make a mistake.
This is not surprising as there is no uncertainty in our
beliefs or the environment. For very small prior beliefs this
bound grows larger which is again intuitive as the contribution
of the true environment to the correct prediction will
initially be small. However as long as the prior is non zero,
Bayes mixture performs well.

The seconds result shows that Bayes mixture will eventually
predict arbitrarily many sequential observations correctly with
probability approaching one. This means it is also an excellent
multi-step look-ahead predictor.

\paradot{Non-deterministic}
In non-deterministic environments there is always uncertainty
about the observation sequence so we need to generalize our
criterion for good performance. At each step the true
environment is going to produce each observation with some
probability, so ideally we want to predict each observation
with this same probability. Therefore, in order to perform
well, we want the distribution given by Bayes mixture to
converge to the true distribution. To analyze this convergence
a notion of distance between the two predictive distributions
is required. For this we use the squared Hellinger distance.
\[
  h_t(x_{<t}) \;:=\; \sum_{a\in\X}\left(\sqrt{\xi(a|x_{<t})}-\sqrt{\mu(a|x_{<t})}\right)^2
\]

This distance is dependent on the previously observed data
$x_{<t}$ because the distributions given by the environments
are also dependent on this data. Intuitively two distributions
are the same when they give equal probabilities to all possible
observations, which is exactly the requirement for the
Hellinger distance to be zero.

Even with this concept of distance the stochastic nature of the
true environment makes mathematical guarantees difficult. For
example a very unlikely sequence of observations may occur
which causes Bayes mixture to be a lot further from the true
distribution than we would expect. Because of this, results are
given in terms of expectations.
For example imagine a coin is flipped one hundred times and you
are given the choice to receive $\$1$ for every heads and $\$0$
for every tails or $\$0.30$ every flip regardless. In the first
case the only guarantee is that you will receive between $\$0$
and $\$100$, however you know that the expectation is $\$50$
which is greater than the certain $\$30$ you would receive in
the second case. Therefore expectation allows us to make
decisions that will be beneficial in the long run.

It was shown in \cite{Hutter:03spupper,Hutter:04uaibook} that
\[
  \sum_{t=1}^\infty \E[h_t] \;\leq\; \ln(w_\mu^{-1}) \;<\; \infty
\]
As before the upper bound is a constant dependent on the prior
belief in the true environment and the intuition is the same.
This result implies that $\xi(x_t|x_{<t})$ will rapidly
converge to $\mu(x_t|x_{<t})$ with probability one as
$t\to\infty$. In a stochastic environment ``with probability
one'' is usually the strongest guarantee that can be made so
this is a strong result. Roughly it states that the expected
number of errors will be finite and Bayes mix will eventually
be the same as the true predictive distribution.

\subsection{Bayesian Decisions}

We have seen that Bayes mixture acts as an excellent predictor
and converges rapidly to the true environment. It is therefore
not surprising that making decisions based on this predictor
will result in excellent behavior. When making decisions
however we are not concerned with the accuracy of our decisions
but rather with the resultant loss or gain. We want to make the
decision that will maximize our expected reward or minimize
our expected loss.

In general, not all wrong predictions are equally bad. When
predicting the rise and fall of stocks for example, a
prediction that is off by only a fraction of a cent is
probably still very useful while a prediction that is off by a
few dollars may be hugely costly. As long as this loss is
bounded we can normalize it to lie in the interval $[0,1]$.
Formally let $\Loss(x_t,y_t)\in[0,1]$ be the received loss when
$y_t$ has been predicted and $x_t$ was the correct observation.

Given this loss function the optimal predictor $\Lambda_\rho$
for environment $\rho$ after seeing observations $x_{<t}$ is
defined as the prediction or decision or action $y_t$ that
minimizes the $\rho$-expected loss. This is the action that we
expect to be least bad according to environment $\rho$.
\[
  y_t^{\Lambda_\rho}(x_{<t}) \;:=\; \arg\min_{y_t}\sum_{x_t}\rho(x_t|x_{<t})\Loss(x_t,y_t)
\]
It should be noted that this optimal predictor may not give the
prediction that is most likely. Imagine we have some test $T$
for cancer. The test result for a patient shows that there is a
$10\%$ chance that the patient has cancer. In other words we
can consider $T$ to be the distribution where
$T(positive|patient)=0.1$. The loss incurred by not predicting
cancer given that the patient does have cancer is $1$ (after
normalization), if the patient doesn't have cancer the loss is
$0$. On the other hand the loss incurred by predicting cancer
if the patient does have cancer is $0,$ while if the patient
doesn't have cancer the loss is a nominal $0.01$ for premature
treatment or further testing.

Given these values the $T$-expected loss of predicting cancer is
\[
  T(+|patient)\Loss(+,+) \;+\; T(-|patient)\Loss(-,+) \;=\; 0.1\times 0 + (1-0.1)\times 0.01 \;=\; 0.009
\]
The $T$-expected loss of not predicting cancer is
\[
  T(+|patient)\Loss(+,-) \;+\; T(-|patient)\Loss(-,-) \;=\; 0.1\times 1 + (1-0.1)\times 0 \;=\; 0.01
\]
Therefore $\Lambda_T$, the optimal predictor for $T$, would
choose to predict cancer even though there's only a $10\%$
likelihood, because it minimized the $T$-expected loss.

Given this optimal predictor, the expected instantaneous loss
at time $t$ and the total expected loss from the first $n$
predictions are defined as follows.
\[
  loss_t^{\Lambda_\rho} \;:=\; \E[\Loss(x_t,y_t^{\Lambda_\rho})] \qmbox{ and }
  Loss_n^{\Lambda_\rho} \;:=\; \sum_{t=1}^n\E[\Loss(x_t,y_t^{\Lambda_\rho})]
\]
Obviously the best predictor possible is the optimal predictor
for the true environment $\Lambda_\mu$, however as $\mu$ is
generally unknown, the best available option is the optimal
predictor $\Lambda_\xi$ for Bayes mixture $\xi$ for which the
following result holds:
\[
  \left(\sqrt{Loss_n^{\Lambda_\xi}}-\sqrt{Loss_n^{\Lambda_\mu}}\right)^2
  \;\leq\; \sum_{t=1}^n \E\!\left[\Big(\sqrt{loss_t^{\Lambda_\xi}}-\sqrt{loss_t^{\Lambda_\mu}}\;\Big)^2\right]
  \;\leq\; 2\ln(w_\mu^{-1}) \;<\; \infty
\]
This means that the squared difference between the square roots
of the total expected losses for $\xi$ and $\mu$ is also
bounded by a constant dependant on our initial belief. This
result demonstrates that from a decision theoretic perspective
the Bayesian mixture as a predictor performs excellently
because it suffers loss only slightly larger than the minimal
loss possible. The bound also implies that the instantaneous
$loss_t^{\Lambda_\xi}$ of Bayes-optimal predictor $\Lambda_\xi$
converges to the best possible $loss_t^{\Lambda_\mu}$ of the
informed predictor $\Lambda_\mu$. In fact one can show that if
a predictor performs better than $\Lambda_\xi$ in any
particular environment then it must perform worse in another
environment. This is referred to as Pareto optimal in
\cite{Hutter:03optisp}.

\subsection{Continuous Environment Classes}\label{sec:cont}

Although the results above were proved assuming that the model class is
countable, analogous results hold for the case that the model
class $\M$ is uncountable such as continuous parameter classes.
For a continuous $\M$ the Bayesian mixture must be defined by
integrating over $\M$.
\[
  \xi(x) \;=\; \int_{\nu\in\M}\nu(x)w(\nu) d\nu
\]
where $w(\nu)$ is (now) a prior probability {\em density} over
$\nu\in\M$. One problem with this is that the dominance
$\xi(x)\ge w(\mu)\mu(x)$ is no longer valid since the prior
probability (not the density) is zero for any single point. To
avoid this problem the Bayesian mixture is instead shown to
dominate the integral over a small vicinity around the true
environment $\mu$. By making some weak assumptions about the
smoothness of the parametric model class $\M$, a weaker type of
dominance makes it possible to prove the following
\cite{Clarke:90,Hutter:03optisp}:
\[
  \sum_{t=1}^n\E[h_t] \;\leq\; \ln(w(\mu)^{-1})+O(\log(n))
\]
This shows that even for a continuous $\M$ we get a similar
bound.
The added logarithmic term means that the sum to $n$ of the
expected Hellinger distance is no longer bounded but grows very
slowly. This is still enough to show that the distribution
given by $\xi$ deviates from the true distribution $\mu$
extremely seldom.
The main point is that the effectiveness of the Bayesian
framework is not significantly impaired by using a continuous
class of environments.

\subsection{Choosing the Model Class}

The above results demonstrate that the Bayesian framework is
highly effective and essentially optimal given the available
information. Unfortunately the operation and performance of
this framework is sensitive to the initial choice of hypothesis
class and prior. As long as they are non zero, the chosen priors
will not affect the asymptotic performance of the Bayesian
mixture as the observations eventually wash out this initial
belief value. However in short-term applications they can have
a significant impact.

The only restriction on the hypothesis class is that it must
contain the true environment. This means any hypothesis that
may be true, however unlikely, should be considered. On the
other hand, having unnecessarily cumbersome classes will affect
the prior values as they must sum to one. This means adding
unnecessary hypotheses will subtract from the priors of
relevant hypotheses. Since the bound of $\ln(w_\mu^{-1})$ is
proportional to the log inverse prior, having unnecessarily
small priors leads to a high error bound which may affect
short-term performance.

For these reasons, the general guideline is to choose the
smallest model class that will contain the true environment
and priors that best reflect a rational a-priori belief in
each of these environments. If no prior information is
available then these priors should reflect this lack of
knowledge by being neutral or objective. In the case of
universal induction however, there is essentially no thinkable
hypothesis we can disregard, so we require a very large model
class.

How to assign reasonable priors over model classes in general
and the model class of essentially all possible explanations in
particular is at the heart of the induction problem. We devote
a whole own section to this intricate issue.

\section{History}\label{cha:history}

The history of the induction problem goes back to ancient times
and is intimately tied to the history of science as a whole.
The induction principle is at the core of how we understand and
interact with our world and it is therefore not surprising that
it was a topic of interest for numerous philosophers and
scientists throughout history, even before the term induction
was used or properly defined. It is however surprising that a
formal understanding of induction is not given greater emphasis
in education when it is, at least implicitly, of fundamental
importance and relevance to all of science. In the following we
will look at some of the most important historical
contributions to inductive reasoning, including recent attempts
at formalizing induction. We will also examine some of the
major problems that plagued these attempts and later these will
be re-examined in the context of universal induction to
illustrate how Solomonoff succeeds where others have failed.

\subsection{Epicurus}

Some of the earliest writings on inductive reasoning are
attributed to the ancient Greek philosopher Epicurus born
roughly 341BC. Epicurus founded the philosophical school of
Epicureanism which taught that observation and analogy were the
two processes by which all knowledge was created. Epicurus's
most relevant teaching in regard to inductive reasoning is his
principle of multiple explanations. This principle states that
{\em ``if more than one theory is consistent with the data,
keep them all''}. Epicurus believed that if two theories
explained some observed phenomenon equally well it would be
unscientific to choose one over the other. Instead both
theories should be kept as long as they both remain consistent
with the data.

This principle was illustrated by one of his followers with the
example of observing a dead body lying far away \cite{Li:08}.
In reasoning about the cause of death we may list all possible
causes such as poison, disease, attack etc, in order to include
the one true cause of death. Although we could reason that one
of these must be the correct cause there is no way of
establishing the true cause conclusively without further
examination and therefore we must maintain a list of possible
causes.

Similar reasoning is used in statistics to derive the principle
of indifference which assigns equal prior probabilities to all
models when there is no reason to initially prefer one over any
other. Although there is clearly a certain validity in
Epicurus's reasoning, it seems unsatisfactory to believe
equally in any hypothesis that accounts for some observed
phenomenon.\textbf{ }To see this consider that you have just
looked at your watch and it is showing the time as $1$pm. It
seems reasonable that you should therefore believe in the
hypothesis that the time is in fact $1$pm and your watch is on
time. But the hypothesis that it is actually $3$pm and your
watch is two hours slow also explains the observations. It is
also possible that a friend set your watch forward three hours
as a joke and it is only $10$am. In fact it is possible to come
up with ever more implausible scenarios which would equally
account for your watch currently showing $1$pm and which would
therefore, according to Epicurus, deserve equal consideration.
So why then do we maintain a strong believe that our watch is
correct and the time is actually $1$pm? If we were entirely
true to Epicurus's principle then a watch would have no use to
us at all since any time would be equally possible regardless
of the time shown. It is clear that our belief in a hypothesis
is directly related to its plausibility and it is this idea of
plausibility which we will further investigate.

Another problem with the principle of indifference is that it
says nothing about how we should choose between conflicting
predictions given by the various consistent models. Since our
primary concern is making good predictions this is a serious
issue.

\subsection{Sextus Empiricus and David Hume}\label{sec:Hume}

Sextus Empiricus was a philosopher born in $160$ AD who gives
one of the first accounts of inductive skepticism. He wrote

{\em When they propose to establish the universal from the
particulars by means of induction, they will effect this by a
review of either all or some of the particulars. But if they
review some, the induction will be insecure, since some of the
particulars omitted in the induction may contravene the
universal; while if they are to review all, they will be
toiling at the impossible, since the particulars are infinite
and indefinite.} \cite{Sextus:33}

This remains the simplest and most intuitive criticism of
universal generalizations. Put simply it states that no
universal generalization can ever be rigorously proven since it
is always possible that an exception will be observed that
will contradict this generalization. There is no flaw in this
reasoning, however to believe that this argument invalidates
induction is to misunderstand inductive reasoning. The argument
does demonstrate that our belief in any universal
generalization should never be $100\%$, but this is widely
accepted and does not hinder the formalization of an inductive
framework. It does however give a sensible criterion which can
be used to test the validity of an inductive method.

Sextus also gave an argument that resembles a better known
argument due to Hume. Sextus wrote {\em ``Those who claim for
themselves to judge the truth are bound to possess a criterion
of truth. This criterion, then, either is without a judge's
approval or has been approved. But if it is without approval,
whence comes it that it is truthworthy? For no matter of
dispute is to be trusted without judging. And, if it has been
approved, that which approves it, in turn, either has been
approved or has not been approved, and so on ad infinitum.''}
 \cite{Sextus:33}

Hume's argument was that induction can not be justified because
the only justification that can be given is inductive and
hence the reasoning becomes circular. Although Hume and
Empiricus reach the same conclusion, that induction can never
be verified and is therefore inherently unreliable, they differ
greatly in how they treat this conclusion.

Sextus believed that since there is no way affirming or denying
any belief, we must give up any judgement about beliefs in
order to attain peace of mind \cite{Annas:00}. It is worth
mentioning that although he was a skeptic of induction,
Empiricus' philosophy was in many respects similar to the
school of Bayesianism. Bayesians would agree that we should not
hold any belief as certain (probability $1$) or deny it
entirely (probability $0$; cf.\ the confirmation problem in
Subsection~\ref{sec:confirm}), apart from logical
tautologies or contradictions, respectively. Empiricus also
believed that an objective truth of reality was unknowable and
instead we can only be sure of our own subjective experiences.
This is similar to the subjective interpretation of
probability.

Hume on the other hand admits that using inductive inference,
or at least reasoning by analogy, is an inevitable part of
being human. He states {\em ``having found, in many instances,
that any two kinds of objects -- flame and heat, snow and cold
-- have always been conjoined together; if flame or snow be
presented anew to the senses, the mind is carried by custom to
expect heat or cold, and to believe that such a quality does
exist, and will discover itself upon a nearer approach. This
belief is the necessary result of placing the mind in such
circumstances''} \cite{Hume:1739}. Hume therefore concedes that
although it can not be verified, induction is essential to our
nature.

\subsection{William of Ockham}

The most important concept for inductive reasoning was famously
posited by William of Ockham (or Occam) although the concept
can not really be attributed to any person as it is simply an
aspect of human nature. In its original form Occam's razor is
stated as {\em ``A plurality should only be postulated if there
is some good reason, experience, or infallible authority for
it''}. A common interpretation is {\em ``keep the simplest
theory consistent with the observations''}. This principle is
apparent in all scientific inquiry and in our day to day
experience.

It is important to understand that a disposition towards
simpler, or more plausible, explanations is simply common sense
and should be entirely uncontroversial. For every observation
we make on a day to day basis there is a multitude of possible
explanations which we disregard because they are far too
complex or unnecessarily convoluted to be plausible.

This was made clear in the example given above which made the
point that even though there are various ways of accounting for
a watch showing a certain time we remain convinced that it is
the correct time. This is because the watch being correct is by
far the simplest explanation. When Fred walks past some houses
on a street numbered $1$, $2$, $3$ respectively it would be
natural for him to belief they are numbered in the standard
manner however they may just as well be numbered according to
the prime numbers which would have the natural continuation $5$
rather than $4$.

If you observe a street sign when searching for a house you
have never visited before it is natural to assume it is correct
rather than that it has been switched. It is even conceivable
that you are witness to a vast hoax or conspiracy and all the
signs you have seen have been changed. Perhaps the map you hold
has itself been altered or you are coincidentally holding the
one section that was misprinted in this edition or which
Google maps got wrong. It is difficult to provide intuitive
examples that do not seem entirely absurd but this is actually
the point in some sense. A disposition towards simplicity is
not only common sense, it is actually necessary for functioning
normally in our world which overloads us with huge amounts
information and countless conceivable explanations. People who
place too much belief in unnecessarily complex ideas are often
the ones labeled paranoid or illogical.

Occam's razor is an elegant and concise formulation of this
natural disposition, however it is still too vague to use
formally. Both `simplest' and `consistent' require precise
definitions but this is difficult due to their inherent
subjectivity. It is also worth noting that even two people
observing the same phenomenon with precisely the same
interpretation of Occam's razor may draw different conclusions
due to their past history. Generally speaking the observations
referred to in Occam's razor are from a specific experiment or
phenomenon but philosophically they can be thought of as all
the observations that make up an individual's life and are
therefore inevitably unique for each person. As a trivial
example imagine Scott was walking with Fred down the street as
he had seen the numbers $1$, $2$, $3$ on the houses but Scott
had previously observed that the next house on the street was
numbered $5$. Scott's beliefs about the continuation of the
sequence would then be different to Fred's. He would be more
inclined than Fred to believe the houses were numbered
according to the primes. Having said this, Scott may still have
a higher belief that the houses are numbered normally with the
$4$ absent for some unknown reason. This is because he has more
evidence for the prime ordering than Fred but given his
previous experience with house orderings the idea of a street
with a prime ordering still seems more complex and hence less
plausible than an explanation for the missing $4$.

Few debate the validity of Occam's razor, however it is exactly
its subjective vague nature which has made it difficult to
formalize. Various approaches to machine learning successfully
apply a formal version of Occam's razor relevant to the problem.
For example when fitting a polynomial to some unknown function
based on some noisy sample data we use a regularization
term to avoid over-fitting. This means that the chosen solution
is a trade off between minimizing the error and minimizing the
complexity of the polynomial. This is not simply for aesthetic
purposes; a polynomial that is chosen only to minimize the
error will generally be far from the function that generated
the data and therefore of little use for prediction.
Unfortunately these methods are problem specific. Formalizing
universal inductive reasoning requires a formal and universal
measure of simplicity. As we will see this is exactly what
Kolmogorov provides.

\subsection{Pierre-Simon Laplace and the Rule of Succession}

Laplace's most famous contribution to inductive inference is
his somewhat controversial rule of succession. For i.i.d.\ data
where the outcome of each trial is either a success or a
failure, this rule gives an estimation of the probability of
success on the next trial. This is almost the same problem
that Bayes formulated at the beginning of his essay.

Let $s$ be the number of successes, $f$ be the number of
failures and $n=s+f$ be the total number of trials, which is
recorded in the binary string $x$. The length of $x$ is the
number of trials and each bit is either a $1$ if the
corresponding trial was a success or a $0$ if it was a failure.
For example if we have had $n=4$ trials that were success,
success, failure, success respectively then $x=1101$, $s=3$, and
$f=1$. The rule of succession states that the probability of
success on the next trial is
\[
  P(x_{n+1}=success=1|x_{1:n}) \;=\; \frac{s+1}{n+2}
\]
The validity of this rule has been questioned but it follows
directly from applying the Bayesian framework with a uniform
prior and i.i.d.\ environment assumed. The derivation
actually provides an informative illustration of how the
Bayesian framework can be applied in practice. To see this,
consider some stationary and independent experiments whose
outcome we can categorize as either a success or a failure
every time it is run, e.g.\ flipping a (biased) coin with heads
being a success. In this case our model class
$\M=\{\t|\t\in[0,1]\}$ is the set of possible probabilities the
experiment may give to success on a single trial. Let
$\t_\mu\in[0,1]$ be the true probability of success on a single
trial. Since the experimental trials are stationary and
independent, $\t_\mu$ remains constant, although our belief
about $\t_\mu$ changes.

According to the definition of conditional probability we have
$P(x_{n+1}=1|x_{1:n})=P(x_{1:n}1)/P(x_{1:n})$ where $x_{1:n}1$
is the string $x_{1:n}$ with $1$ appended at the end. The
probability of any particular sequence $x$ of failures and
successes clearly depends on $\t_\mu$ and is given by
$P(x|\t_\mu)=\t_\mu^s(1-\t_\mu)^f$. The intuition behind this
should be clear. Imagine you have a biased coin which gives
heads with probability $\t$ and hence tails with probability
$(1-\t$). The probability of throwing heads $s$ times is
$\t^s$ and the probability of throwing tails $f$ times in a
row is $(1-\t)^f$. Therefore, since the throws are i.i.d.,
the $\t$-probability of any sequence of throws with $s$ heads and
$f$ tails is $\t^s(1-\t)^f$. For a regular, unbiased coin,
the probability of heads is $\t=0.5=(1-\t)$ and therefore the
probability of a sequence is $0.5^{s+f}$. This means it depends
only on the total number of throws.

We are now interested in the probability of some observation
sequence $x$. Since $\t$ is unknown we estimate the true
probability using Bayes mixture which represents our subjective
probability. This involves integrating over our prior belief
density $w(\t)=P(\t)$ to give
\[
  \xi(x) \;=\; P(x) \;=\; \int_0^1P(x|\t)w(\t)d\t
\]
Note that we can not sum because the model class $\M$ is
continuous and hence uncountable. Since we assume the prior
distribution to be uniform and proper it must satisfy the
following constraints
\[
  \int_0^1w(\t)d\t \;=\; 1 \qmbox{and}
  w(\t) \;=\; w(\t') \mbox{ for all } \t \mbox{ and } \t'\in\M
\]
This results in the density, $w(\t)=1$ $\forall\t\in[0,1]$.
Therefore
\[
P(x) \;=\; \int_0^1P(x|\t)d\t
     \;=\; \int_0^1\t^s(1-\t)^f d\t
     \;=\; \frac{s!f!}{(s+f+1)!}
\]
The final equality is a known property of the Beta
function. To find the conditional probability of success given
this sequence we need to consider the sequence $x_{1:n}$ with
another success appended at the end. This is denoted
$x_{1:n}1$. The probability of this sequence $P(x_{1:n}1)$
follows analogously from above since this sequence contains the
same number of failures as $x_{1:n}$ plus one more success.
Therefore
\[
P(x_{n+1}=1|x_{1:n}) \;=\; \frac{P(x_{1:n}1)}{P(x_{1:n})}
 \;=\; \frac{\frac{(s+1)!f!}{(s+1+f+1)!}}{\frac{s!f!}{(s+f+1)!}}
 \;=\; \frac{s+1}{s+f+2}
 \;=\; \frac{s+1}{n+2}
\]
The controversial and perhaps regretful example Laplace
originally used to illustrate this rule was the probability
that the sun will rise tomorrow given that it has risen in the
past. Laplace believed that Earth was $5000$ years old and
hence that the sun had risen $1826213$ times previously.
Therefore by his rule the probability that the sun will rise
tomorrow is $\frac{1826214}{1826215}$ or equivalently the
probability that it won't rise is $\frs{1}{1826215}$. In his
original statement of the problem Laplace appended the example
immediately with a note that this is only applicable if we knew
nothing else about the mechanism of the sun rising, but since
we know a lot about this mechanism the probability would be far
greater. Unfortunately many ignored this accompanying
explanation and claimed that the rule was invalid because this
estimate for the sun rising was simply absurd. One counter
example claimed that by the same reasoning the probability that
a $10$ year old child will live another year is $\frac{11}{12}$
compared with an $80$ year old man having probability
$\frac{81}{82}$ \cite{Zabell:89} even though this clearly is a
case where previous knowledge and lack of independence make the
rule inapplicable.

\subsection{Confirmation Problem}\label{sec:confirm}

The confirmation problem involves the inability to confirm
some particular hypothesis in the hypothesis class regardless
of the evidence we observe for it. Usually the hypotheses
considered when examining this problem are universal
generalizations. Dealing with universal generalizations remains
one of the most persistent challenges for systems of inductive
logic.

As Empiricus argued, verifying any universal generalization is
difficult because regardless of how many instances of something
we observe, it is always possible that some unobserved instance
will contradict any conclusions we may have drawn thus far.
Therefore, finding a system that gives complete certainty in a
universal generalization cannot be the actual goal since we
know that this could not be justified unless we have observed
every possible instance of the object about which the
generalization was made. The problem is in finding a system
that agrees with our intuition in all aspects of inductive
logic. However satisfactory behavior regarding the confirmation
of universal generalizations has evaded most proposed systems.

It should be noted that in this section and throughout the
article we have used the term confirmation as it appears
commonly in the literature \cite{Maher:04}. The meaning in this
context is similar to what an unfamiliar reader may associate
more closely with the term {\em `supports'}. To make this
clear, some evidence $E$ is said to confirm a hypothesis $H$ if
the posterior probability of $H$ given $E$ is greater than
before $E$ was observed. Unfortunately, as is illustrated in
the next subsection, this definition is unsatisfactory in
certain circumstances. An arbitrarily small increment in belief
may not deserve to be labeled as confirmation, and so we call
it weak confirmation. The confirmation problem is illustrated
here using the above rule of succession.

When applicable, Laplace's rule of succession seems to produce
a reasonable estimate. It converges to the relative frequency,
it is defined before any observations have been seen (for
$s=f=0$) and it is symmetric. It also isn't over-confident
meaning that it never assigns probability 1 to an observation.
This satisfies Epicurus's argument that no induction can ever
be certain. Unfortunately there are some significant
draw-backs, namely the zero prior problem. This zero prior
problem is a specific instance of the more general confirmation
problem.

The zero prior problem occurs because of the prior factor
present in Bayes rule. For some hypothesis $H$ and evidence $E$
Bayes rule states $P(H|E)=P(E|H)P(H)/P(E)$. Therefore it is
clear that if $P(H)=0$ then regardless of the evidence $E$ our
posterior evidence $P(H|E)$ must remain identically zero. This
is why any hypothesis which is possible, no matter how
unlikely, must be assigned a non-zero prior. When approximating
the probability of success in the biased coin-flip example
above, the hypothesis $H$ corresponds to a particular $\t$, and
the evidence $E$ is the observation sequence $x$. Although the
densities $w(\t)$ are non zero for all $\t$, the probability of
any particular $\t$ is $P(\t=c)=\int_{c}^{c}w(\t)d\t=0$. Any
proper density function must have zero mass at any point. This
means that for any $\t$, the posterior $P(\t|x)=0$ no matter
what the observation sequence. Imagine we are observing the
color of ravens and $\t$ is the percentage of ravens that are
black. The hypothesis ``All ravens are black'' therefore might
be associated with $\t=1$, but even after observing one million
black ravens and no non-black ravens $P(\t=1|x)=0$, which means
we are still certain that not all ravens are black:
$P(\t<1|x)=1$. This is clearly a problem.

If we instead consider the composite or partial hypothesis
$\t_p=\{\t|\t\in(1-\varepsilon,1]\}$ for any arbitrarily small
$\varepsilon$, then $P(\t|x)$ converges to $1$ as the number of
observed black ravens increases. This is called a soft
hypothesis and intuitively it is the hypothesis that the
percentage of black ravens is $1$ or very close to $1$. The
reason our belief in this hypothesis can converge to $1$ is
that the probability is now the integral over a small interval
which has a-priori non-zero mass $P(\t_p)=\varepsilon>0$ and
a-posteriori asymptotically all mass $P(\t_p|1^n)\to 1$.

Instead of $\t=1$ it is also possible to formulate the
hypothesis ``all ravens are black'' as the observation sequence
of an infinite number of black ravens, i.e.\ $H'=x=1^\infty$
where a $1$ is a black raven. This purely observational
interpretation might be considered philosophically more
appropriate since it considers only observable data rather than
an unobservable parameter.  However the same problem occurs. If
$x_{1:n}=1^n$ is a sequence of $n$ black ravens, then
$P(x_{1:n}) = n!/(n+1)! = \frs{1}{n+1}$. Therefore
\[
  P(1^{k}|1^n) \;=\; \frac{P(1^{n+k})}{P(1^n)} \;=\; \frac{n}{n+k}
\]
This means that for any finite $k$ our belief in the hypothesis
that we will observe $k$ more black ravens converges to $1$ as
the number of observed ravens $n$ tends to infinity, which is
not surprising and conforms to intuition. Once we have seen
$1000$ black ravens we strongly expect that we will observe
another $10$ black ravens. However for the above hypothesis of
``all ravens are black'' $k$ is infinite and the probability
$P(1^{k=\infty}|1^n)$ will be zero for any number $n$ of
observed ravens. By making the reasonable assumption that the
population of ravens is finite, and therefore that $k$ is
finite, we may expect to fix the problem. This is the approach
taken by Maher \cite{Maher:04}. However it still leads to
unacceptable results which we examine further in the next
subsection.

Since both forms of the universal generalization fail to be
confirmed by the rule of succession, there seem to be only two
reasonable options.
We can simply accept that hypotheses corresponding to exact
values of $\t$ can not be confirmed, so instead soft hypotheses
corresponding to small intervals or neighborhoods must be used.
While we can successfully reason about soft hypotheses, we
still have to decide what to do with the universal hypotheses.
We would somehow have to forbid assigning probabilities to
all-quantified statements. Assigning probability zero to them
is not a solution, since this implies that we are certain that
everything has exceptions, which is unreasonable. We can also
not be certain about their truth or falsity. Bare any semantic,
we could equally well eliminate them from our language. So
focussing on soft hypotheses results in a language that either
does not include sentences like ``all ravens are black'' or if
they exist have no meaning. This makes the soft hypothesis
approach at best inelegant and impractical if not infeasible.

The other solution is to assign a non-zero weight to the point
$\t=1$ \cite{Zabell:89}. This point mass results in an improper probability
density however it does solve the confirmation problem. One
such improper distribution is a 50:50 mixture of a uniform
distribution with a point mass at 1. Mathematically we consider
the distribution function $P(\t\ge a)=1-\frac{1}{2}a$ with
$a\in[0,1]$, which gives $P(\t=1)=\frs{1}{2}$. Using this
approach results in the following Bayesian mixture
distribution, again with $s$ successes, $f$ failures and
$n=s+f$ trials:
\[
  \xi(x_{1:n}) \;=\; \frac{1}{2}\left(\frac{s!f!}{(n+1)!}+\delta_{s,n}\right) \qmbox{where} \delta_{s,n} =
  \begin{cases}
    1 & \mbox{if }s=n\\
    0 & \mbox{otherwise}
  \end{cases}
\]
Therefore, if all observations are successes, or black ravens,
the Bayesian mixture gives $\xi(1^n) =
\frac{1}{2}(\frac{n!0!}{(n+1)!}+1) =
\frac{1}{2}\cdot\frac{n+2}{n+1}$, which is much larger than the
$\xi(1^n)=\frs{1}{n+1}$ given by the uniform prior. Because of
this both the observational hypothesis $H':=(x=1^\infty)$ and
the physical hypothesis $\t=1$ can be confirmed by the
observation of a reasonable number of black ravens. Formally,
the conditional distribution of seeing $k$ black ravens after
seeing $n$ black ravens is given by
\[
  P(1^{k}|1^n) \;=\; \xi(1^{k}|1^n)
  \;=\; \frac{\xi(1^{n+k})}{\xi(1^n)}
  \;=\; \frac{n+k+2}{n+k+1}\cdot\frac{n+1}{n+2}
\]
\[
  \mbox{Therefore}\quad P(H'|1^n) \;=\; P(1^\infty|1^n)
  \;=\; \lim_{k\to\infty}P(1^{k}|1^n) \;=\; \frac{n+1}{n+2}
\]
and hence the observational hypothesis $H'$ is confirmed with
each new observation. Our confidence in the hypothesis that all
ravens are black after having observed 100 black ravens is
about $99\%$.
The first line also shows confirmation occurs for any finite
population $k$. As we would expect the physical hypothesis
similarly gets confirmed with $P(\t=1|1^n)=\frac{n+1}{n+2}$.
The new prior also has the property that once a non-black raven
is observed, the posterior Bayesian distribution becomes the
same as it would have been if a uniform prior had been assumed
from the start, since $\delta_{s,n}=0$ in this case.

So far we have considered a binary alphabet, but the idea of
assigning prior point masses has a natural generalization to
general finite alphabet. For instance if we instead consider
the percentage of black, white and colored ravens, the results
remain analogous.

It is immediately clear that the chosen ``improper density''
solution is biased towards universal generalizations, in this
case to the hypothesis ``all ravens are black''. The question
is then why not design the density to also be able to confirm
``no ravens are black'', or ``exactly half the ravens are
black''? It would be possible to assign a point mass to each of
these values of $\t$ but then why only these values? These
values correspond to hypotheses that seem more reasonable or
more likely and therefore which we want to be able to confirm.
But ideally we want to be able to confirm any reasonable
hypothesis, so the question becomes which points correspond to
reasonable hypotheses?

It seems that we are intuitively biased towards hypotheses
corresponding to simpler values such as rational numbers but we
can argue that significant irrational fractions such as $1/\pi$
are also very reasonable. Deciding where to draw the line is
clearly problematic but the universal prior which is described
later provides a promising solution. It assigns non-zero
probability to any computable number, and the class of
computable numbers certainly contains any reasonable values
$\t$. A non-computable $\t$ corresponds to a non-computable
hypothesis, which are usually not considered (outside of
mathematics). It should also be noted that even if $\t$ is
incomputable, there are always arbitrarily close values which
are computable and hence can be confirmed. Formally this means
that the set of computable numbers is dense in the real
numbers. The universal prior can therefore be seen as a logical
extension of the above method for solving the confirmation
problem.

Since this class of computable values is infinite it may be
asked why we don't go one step further and simply assign every
value a non-zero point mass. The reason is that it is not
mathematically possible. The reason comes down to the
difference between countably infinite and uncountably infinite.
Without going into depth consider the infinite sum
$\sum_{n=1}^\infty 2^{-n}=1$. The property of creating an
infinite sum that gives a finite value is only possible for
countably infinite sums and since the set of real numbers in
the interval $[0,1]$ is uncountably infinite it is not possible
to assign values that form an everywhere non-zero prior.

\subsection{Patrick Maher does not Capture the Logic of Confirmation}

In his paper ``probability captures the logic of scientific
confirmation'' \cite{Maher:04} Patrick Maher attempts to show
that by assuming only the axioms of probability it is possible
to define a predicate that captures in a precise and
intuitively correct manner the concept of confirmation. Maher
chooses to use a conditional set of probability axioms based on
that of von Wright, presumably for convenience.

Maher's definition of confirmation is
\[
  \qmbox{\it Definition:} C(H,E,D) \qmbox{iff} P(H|E.D)>P(H|D)
\]
Intuitively meaning that some evidence $E$ confirms a
hypothesis $H$ when the probability of $H$ given $E$ and some
background knowledge $D$ is greater than the probability of $H$
given $D$ alone. It is generally agreed that any attempt to
define confirmation must consider background knowledge. This is
illustrated in the following example by I.J.~Good
\cite{Good:60}.

Suppose our background knowledge is that we live in one of two
universes. In the first there are 100 black ravens, no
non-black ravens and 1 million other birds. In the second there
are 1000 black ravens, 1 white raven and 1 million other birds.
Some bird $a$ is selected at random from all the birds and is
found to be a black raven. It is not hard to see that in this
case the evidence that $a$ is a black raven actually lessens
our belief that `all ravens are black' since it increases the
probability that we are in the second universe where this is
false.

Maher successfully shows that the above definition satisfies
several desirable properties regarding our intuition of
confirmation and scientific practice such as verified
consequences and reasoning by analogy. Unfortunately this
definition fails to satisfactorily solve the problem of
universal generalizations. To illustrate this problem we again
consider confirmation of the universal generalization ``all
ravens are black''. In particular, given that we have observed
$n$ black ravens what is our belief that all ravens are black?

Consider Theorem~9 from Maher's paper. For this example we
assume that $a$ is drawn at random from the population of
ravens and we take the predicate $F(a)$ to mean that $a$ is
black.
\[
  \mbox{\it Theorem~9:~~~If } P(E)>0 \mbox{ then } P(\forall x\, F(x)|E)=0
\]
This means that regardless of the evidence, as long as its
logically consistent, our belief in the universal
generalization $\forall x\, F(x)$ remains zero. This is clearly
a problem since although our belief in this generalization
should not be $100\%$ certain it should be greater than zero as
long as the evidence does not contradict the generalization. In
particular it should be possible to observe some evidence $E$,
such as many $x$ for which $F(x)$ holds, which leads to a
significant posterior belief in this universal generalization.

The reason for this problem, under Maher's construction, is
that the probability that the next observed raven is black
converges to one too slowly. After seeing a long enough
sequence of black ravens our belief that the next one is black
will become arbitrarily close to one but it is the rate of this
convergence that is a problem. Because of this, the
probability that all ravens are black remains zero regardless
of our initial belief. A corollary of this is Maher's
Theorem~10 which, for any logical truth $T$, states
\[
  \mbox{\it Theorem~10:~~~ } \forall n\in\mathbb{N}\;\neg C(\forall x\, F(x),\, F(a_1)...F(a_n),T)
\]
Intuitively this means that there is no evidence that can be said
to confirm a universal generalization. Consider {$F(a_1),...F(a_n)$
to be the evidence $E$ in Theorem~9. Since the posterior belief
in the universal generalization must always remain zero for any evidence
it is clear that this evidence can not increase the belief. Therefore
it can not satisfy Maher's above definition of confirmation.

In observing that the zero probability of universal
generalizations stems from the infinite product in the proof of
Theorem~9, Maher attempts to rectify the problem by considering
only a finite population which he states is sufficient. Even if
we accept the finiteness assumption, the solution he provides
differs dramatically from accepted intuition. Theorem~11 is
where we see the major flaw in Maher's reasoning.
\[
  \mbox{\it Theorem~11:~~~~ } \forall n,N\in\mathbb{N}\;\; C(F(a_1)...F(a_N),F(a_1)...F(a_n),T)
\]
If there are only $N$ ravens in existence then the universal
generalization $\forall x\, F(x)$ is equivalent to $N$
individual observations $F(a_1)...F(a_N)$. In other words, as
long as there is some finite population $N$ of ravens any
observed subset $n$ of ravens confirms the universal
generalization. This is technically correct but we see from the
following numerical example that it is unacceptable. In order
to be fair to Maher the example is constructed similar to his
own numerical example.

Let the population of ravens in the world be $N=1'000'000$ and
the number of observed ravens be $n=1000$. The learning rate is
$\lambda=2$ and we assume the initial belief that some raven is
black to be an optimistic $\gamma_{F}=0.5$. By Maher's
Proposition~19, the degree of belief in the black raven
hypothesis can be computed as follows:
\[
  P(F(a_1)...F(a_N)) \;=\; \prod_{i=0}^{N-1}\frac{i+\lambda\gamma_{F}}{i+\lambda}
  \;=\; \frac{1}{N+1} \;\doteq\; 0.000001
\]
\[
  P(F(a_1)...F(a_n)) \;=\; \prod_{i=0}^{n-1}\frac{i+\lambda\gamma_{F}}{i+\lambda}
  \;=\; \frac{1}{n+1} \;\doteq\; 0.001
\]
\[
  \mbox{Therefore}\quad P(F(a_1)...F(a_N)|F(a_1)...F(a_n))
  \;=\; \frac{P(F(a_1)...F(a_N))}{P(F(a_1)...F(a_n))}
  \;=\; \frac{n+1}{N+1} \;\doteq\; 0.001
\]
This means that after observing 1000 ravens which were all
black our belief in the generalization {\em `all ravens are
black'} is still only $0.1\%$. In other words we are virtually
certain that non-black ravens exist or equivalently that not
all ravens are black. This is a clear contradiction to both
common sense and normal scientific practice and therefore we
must reject Maher's proposed definition. This model of
confirmation is too weak to achieve a reasonable degree of
belief in the black ravens hypothesis. In contrast, in
Section~\ref{sec:BR} we show that Solomonoff exhibits strong
confirmation in the sense that $P$ tends to 1. It may be
believed that this result is due to this particular setup of the
problem, however any continuous prior density and reasonable
parameter values will encounter the same problem. In particular
this includes Maher's more sophisticated model for two binary
properties, which mixes a Laplace/Carnap model for blackness
times one for ravenness with a Laplace/Carnap model where the
properties are combined to a single quaternary property.
Observing a small fraction of black ravens is not sufficient to
believe more in the hypothesis than in its negation, since the
degree of confirmation in Maher's construction is too small.

\subsection{Black Ravens Paradox}

We have used the typical example of observing black ravens to
demonstrate the flaws of both Laplace and Maher in relation to
confirmation but the full `black ravens paradox' is a deeper
problem. It is deeper because even in a system that can confirm
universal hypotheses, it demonstrates a further property that
is highly unintuitive.

The full black ravens paradox is this: It has been seen that
one desirable property of any inductive framework is that the
observation of a black raven confirms the hypothesis that ``all
ravens are black''. More generally we would like to have the
following property for arbitrary predicates $A$ and $B$. The
observation of an object $x$ for which $A(x)$ and
\textbf{$B(x)$ }are true confirms the hypothesis ``all $x$
which are $A$ are also $B$'' or $\forall x$ $A(x)\limpl
B(x)$. This is known as Nicods condition which has been seen as
a highly intuitive property but it is not universally accepted
\cite{Maher:04}. However even if there are particular
situations where it does not hold it is certainly true in the
majority of situations and in these situations the following
problem remains.

The second ingredient to this paradox is the interchangeability
of logically equivalent statements in induction. In particular,
consider two logically equivalent hypotheses $H_1$ and $H_2$.
If some evidence $E$ confirms hypothesis $H_1$ then it
logically follows that $E$ also confirms $H_2$, and vice versa.
But any implication of the form $A\limpl B$ is logically
equivalent to its contrapositive $\neg B\limpl\neg A$.
Therefore, taking the predicate $R(x)$ to mean ``is a raven''
and $B(x)$ to mean ``is black'', this gives the following: The
hypothesis $\forall(x)\, R(x)\limpl B(x)$, or ``all ravens
are black'', is logically equivalent to its contrapositive
$\forall(x)\,\neg B(x)\limpl\neg R(x)$, or ``anything
non-black is a non-raven''.

The fact that any evidence for $\forall(x)\,\neg
B(x)\limpl\neg R(x)$ is also evidence for $\forall(x)\,
R(x)\limpl B(x)$ leads to the following highly unintuitive
result: Any non-black non-raven, such as a white sock or a red
apple confirms the hypothesis that ``All ravens are black''.

This may be seen as evidence that there is a fundamental flaw
in the setup being used here, but on closer examination it is
not entirely absurd. To see this, consider the principle in a
more localized setup. Imagine there is a bucket containing some
finite number of blocks. You know that each of these blocks is
either triangular or square and you also know that each block
is either red or blue. After observing that the first few
blocks you see are square and red you develop the hypothesis
``all square blocks are red''. Following this you observe a
number of blue triangular blocks. According to the above
principle these should confirm your hypothesis since they
confirm the logically equivalent contrapositive, ``All non-red
(blue) blocks are non-square (triangular)''. If the statement
were false then there must exist a counter example in the form
of at least some blue square block. As you observe that a
growing number of the finite amount of blocks are not counter
examples your probability/belief that they exist decreases and
therefore the two equivalent hypotheses should be confirmed.

In this simplified case it is also easier to see the intuitive
connection between the observation of blue triangular blocks
and the hypothesis ``all square blocks are red''. Even if there
were an infinite number of blocks, which means the chance of a
counter example does not obviously diminish, the confirmation
of the hypothesis ``all square blocks are red'' by a blue
triangular block seems reasonable. The reason for this is the
following. If there is an infinite number of objects then there
is always the same infinite number of objects that may be
counter examples, but the longer we go without observing a
counter example the more sure we become that they do not exist.
This human tendency is implicity related to the assumption of
the principle of uniformity of nature which is discussed
briefly later. We expect that eventually the sample we see will
be representative of the entire population and hence if there
are no counter examples in this sample they should be unlikely
in the wider population.

In our real-world example of black ravens we can argue for this
principle analogously. When we see a white sock it is
technically one more item that can no longer be a counter
example to the hypothesis ``all ravens are black''. And
although there may be an incomprehensively huge number of
possible objects in our universe to observe, there is still
only a finite amount of accessible matter and hence a finite
number of objects. But this does not seem to change our strong
intuition that this result is ridiculous. No matter how many
white socks or red apples we observe we don't really increase
our belief that all ravens are black. The solution to this
problem lies in the relative degree of confirmation. The above
result only states that the belief in the hypothesis must
increase after observing either a black raven or a white sock,
it says nothing about the size of this increase. If the size of
the increase is inversely proportional to the proportion of
this object type in the relevant object population then the
result can become quite consistent and intuitive.

Consider again the bucket of blocks. First imagine the number
of square and triangular blocks is the same. In this case
observing a red square block or a blue triangular block should
provide roughly the same degree of confirmation in the
hypothesis ``all square blocks are red''.
Now imagine that only $1\%$ of the blocks are square and $99\%$
are triangular. You have observed $20$ blue triangular blocks
and suddenly you observe a red square block. Intuitively, even
if the blue triangular blocks are confirming the hypothesis
``all red blocks are square'', it seems the observation of a
red square block provides substantially more evidence and hence
a much greater degree of confirmation. The higher the
proportion of blue triangular blocks, the less confirmation
power they have, while the smaller the proportion of blue
blocks, the higher their confirmation power.

This also solves the problem of our intuition regarding black
ravens. Black ravens make up a vanishingly small proportion off
all possible objects, so the observation of a black raven gives
an enormously greater degree of confirmation to ``all ravens
are black'' than a non-black non-raven. So much so that the
observation of a non-black non-raven has a negligible affect on
our belief in the statement.

Unfortunately no formal inductive system has been shown to
formally give this desired result so far. It is believed that
Solomonoff Induction may be able to achieve this result but is
has not been shown rigorously. Later we will argue the case for
Solomonoff induction.

\subsection{Alan Turing}

In 1936 Alan Turing introduced the Turing machine. This
surprisingly simple hypothetical machine turned out to be the
unlikely final ingredient necessary for Solomonoff's induction
scheme as it allows for a universal and essentially objective
measure of simplicity.

Turing's aim was to capture the fundamental building blocks of
how we undertake a task or procedure in a way that was general
enough to describe a solution to any well defined problem. The
final product is very minimal consisting of only a few core
components. A Turing machine has a single work tape of infinite
length which it can read from and write to using some finite
number of symbols. The reading and writing is done by a
read/write head which can only operate on one symbol at a time
before either halting or moving to a neighboring symbol. The
rest of the Turing machine is specific to the task and consists
of the procedural rules. These rules can be represented by
internal states with transitions that depend on what tape
symbol is read and which in turn determine which tape symbol is
written. These states can also be replaced by a look-up table
that store the equivalent information. A comprehensive
understanding of precisely how Turing machines work is not
necessary for the purpose of this article as they are only
dealt with on an abstract level. It is important however to
have an intuitive understanding of their capabilities and
properties.

It turns out that this simple construction is incredibly
powerful. The Church-Turing Thesis states that {\em
``Everything that can be reasonably said to be computed by a
human using a fixed procedure can be computed by a Turing
machine''}. There have been various attempts at defining
precisely what a `fixed procedure' is, however all serious
attempts have turned out to describe an equivalent class of
problems. This class of computable functions or problems is
actually large enough to include essentially any environment or
problem encountered in science. This is because every model we
use is defined by precise rules which can be encoded as an
algorithm on a Turing machine. At a fundamental level every
particle interaction is determined by laws that can be
calculated and hence the outcome of any larger system is
computable. The quantum mechanics model is problematic as it
implies the existence of truly random natural processes but as
long as a Turing machine is given access to a truly random
source of input then even this model can be captured.

Although a Turing machine can be constructed for any computable
task it is far from unique. For every task there is actually an
infinite number of Turing machines that can compute it. For
example there are an infinite number of programs that print
``hello world''.

Strictly speaking Turing machines are hypothetical because of
the requirement of an infinite work tape. Nevertheless we can
think of a Turing machine as a computer with finite memory but
which can be arbitrarily extended as it is required. Then the
analogy of Turing machines and real computers actually becomes
an equivalence. There are actually two valid analogies that
can be drawn, which illustrates an interesting property of
Turing machines. First consider the entire memory of the
computer to be analogous to the work tape of the Turing machine
and the program counter to be the position of the read/write
head. Under this analogy the hardware makes up the procedural
rules that govern how memory is written to and read from.
Secondly consider some program running on this computer. Now
only some of the physical memory corresponds to the work tape
and the memory that holds the program instructions corresponds
to the procedural rules. Not only are both of these analogies
valid, they can both be true at the same time. A program can be
thought of as a Turing machine for a specific task which is
itself encoded in some language (ultimately binary) and a
computer can be thought of as a Turing machine that simulates
these encoded Turing machines.

This ability to create a Turing machine to simulate any other
Turing machine is crucial to Solomonoff's framework. Turing
machines with this property are called Universal Turing
machines and just as with any other task, there is an infinite
number of them corresponding to the infinitely many ways of
encoding a Turing machine as a string.

\subsection{Andrey Kolmogorov}\label{sec:kolmo}

The same Kolmogorov who introduced the by now standard axioms
of probability was also interested in universal notions of
information content in or complexity of objects. Kolmogorov
complexity quantifies the troublesome notion of complexity and
hence also simplicity, which is crucial for a formal
application of Occam's razor. Before looking at Kolmogorov's
formal definition it is useful to review our intuitive
understanding of simplicity.

\paradot{Simplicity}
The idea of simplicity is extremely broad as it can be applied
to any object, model, function or anything that can be clearly
described. It is exactly this idea of a description which is
useful in finding a general definition. Let $A$ be some
arbitrary object, it could be as simple as a coffee mug. Let
$B$ be the same as $A$ except with some added detail or
information such as a word printed on the mug. Now it is
natural to think of $A$ as simpler than $B$ because it takes
longer to precisely describe $B$ than $A$. This idea of
description length turns out to be the most general and
intuitive method for quantifying complexity.

Consider two strings $x$ and $y$ where $x$ is a random thousand
digit number and $y$ is one thousand $9s$ in a row. At first
it may seem these two strings are just as complex as each other
because they each take a thousand digits to describe however
{\em ``one thousand $9s$ in a row''} is also a complete
description of $y$ which only requires twenty five characters.
There are many possible descriptions of any string so a
decision must be made as to which description to associate with
the string's complexity. Since there are always arbitrarily
long descriptions the answer is to take the length of shortest
possible description as the complexity measure. It is clear
then that $y$ is simpler than $x$ since it has a far shorter
description. $x$ was also described with the short sentence
{\em ``a random thousand digit number'' }but this was not a
complete description. There are many numbers that could be a
random thousand digit number but only one number is one
thousand $9s$ in a row. The shortest complete description of
any 1000 digit random string is the string itself, hence about
1000 digits long.

Accepting that the simplicity of an object is given by its
shortest possible description the issue of subjectivity remains
in the choice of description language used. It is clear that
the length of a description may be different in different
languages and in the extreme case an arbitrarily complex string
$c$ can have an arbitrarily short description in a language
constructed specifically for the purpose of describing
$c$.\textbf{ }This problem can be avoided by choosing a single
unbiased language to use for all descriptions.

\paradot{Kolmogorov Complexity}
Kolmogorov's idea was to use Turing machines to formally
address the problem of subjectivity in the choice of
description language. This is because a description of an
object can be thought of as a procedure for producing an
unambiguous encoding of that object. In other words a
description is a program. Coming back to the previous example,
a formal coding of {\em ``one thousand $9s$ in a row''} may be
{\em ``for(i=0;i$<$1000;i++) printf(``9'');''}.

There may of course be shorter descriptions but this at least
gives an upper bound on the shortest description. The random
number $x=01100101...10011$ on the other hand would still have
to be written out entirely which would result in a much longer
shortest description {\em ``printf(``01100101...10011'');''}.
If $x$ could be specified by a shorter description, then it
would contain some structure, so by definition it would not be
random.

By using programs we are again faced with the problem of
choosing the programming language, however all programming
languages are compiled to the native assembly language before
being interpreted by the computer. Assembly (at least for RISC
or Lisp processors) provides a rather unbiased, and surely
universal language. This is now close to Kolmogorov's formal
definition. It is worth noting here that the extreme cases of
languages tailored for a specific description is practically
prevented by using assembly language. Consider the case where
we attempt to `cheat' the system by hard-coding a long complex
string such as $c$ as a simple variable $j$ in some new
programming language. Although {\em ``print $j$''} is now a
simple description of $c$ in this new language, when compiled
to assembler the complexity of $c$ becomes clear since the
assembly code for this program will still need to contain its
full description of the hard-coding of $j$.

A specific program in any language can be thought of as an
encoding of a Turing machine and likewise a Turing machine can
be thought of as a program. A Universal Turing machine can be
used to simulate these encoded Turing machines or programs. This
means that if a program/Turing machine $p$ produces $y$ when
given $x$, then a universal Turing machine will also produce
$y$ when given $x$ and $p$. Since native assembly language can
represent any program it can be thought of as a particular
Universal Turing machine. Therefore, taking the description
with respect to assembly language is essentially the same as
taking the description with respect to this particular
Universal Turing machine.
 Since native assembly is written in binary we
consider the description alphabet to be binary also.

Formally the Kolmogorov complexity of a string $x$ is defined as
\[
  K(x) \;:=\; \min_p\{\length(p):U(p)=x\}
\]
Where $U$ is the Universal reference Turing machine, and
$\length(p)$ is the length of $p$ in binary representation.

In other words, the Kolmogorov complexity of $x$ is the length
of the encoding of the shortest program $p$ that produces $x$
when given as input to the Universal reference Turing machine.

\paradot{Conditional Kolmogorov complexity}
In some cases it is necessary to measure the complexity of an
object or environment relative to some given information. This
is done using the conditional Kolmogorov complexity.
Let $x$ be some string and imagine we want to measure the
complexity of $x$ in relation to some previous knowledge, or
side information, $y$. The conditional Kolmogorov complexity is
defined as follows
\[
  K(x|y) \;:=\; \min_p\{\length(p):U(y,p)=x\}
\]
In other words it is the length of the shortest program to
output $x$ given $y$ as extra input. This means that the
information or structure present in $y$ may be used to shorten
the shortest description of $x$. If $y$ is uninformative or
unrelated to $x$ then $K(x|y)$ will be essentially the same as
$K(x)$. However if $y$ contains a lot of the information
relevant to $x$ then $K(x|y)$ will be significantly smaller. As
an example consider an environment $h=y^n$ that simply repeats
a long complex sequence $y$ over and over. $K(h)$ will
therefore be proportional to the complexity of $y$. If,
however, the side information $z$ contains at least one
iteration of $y$ then it is easy to construct a simple
short program that takes the relevant substring of $y$ and
copies it repeatedly. Therefore $K(h|y)$ will be very small.

\begin{wrapfigure}{r}{0.5\textwidth}
  \vspace{-2ex}
  \includegraphics[width=0.5\textwidth]{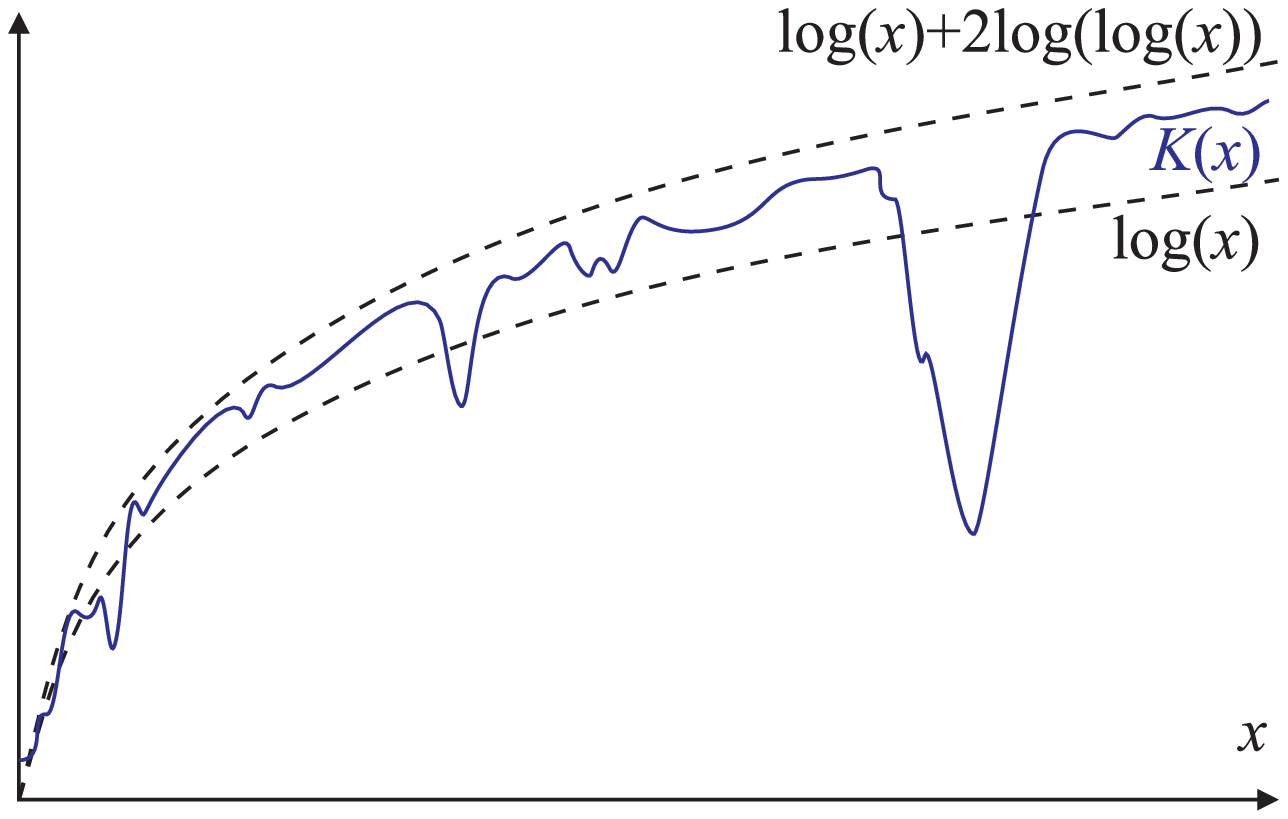}
  \vspace{-5ex}
\end{wrapfigure}
This introduction to Kolmogorov complexity was necessarily
cursory. Kolmogorov complexity possesses many amazing
properties and relations to algorithmic randomness and Shannon
entropy. There are also many variations, and indeed throughout
this work, $K$ stands for the prefix/monotone version if
applied to finite/infinite strings. The differences are
technically important, but are of no concern for us. The
definition of $K$ has natural extensions to other non-string
objects $x$, such as natural numbers and functions, by
requiring $U$ to produce some binary representation of $x$. See
\cite{Hutter:07ait,Li:08} for a list of properties and a
discussion of the graphical sketch of $K$ on the right.

\paradot{Natural Turing Machines}\label{sec:natTuring}
The final issue is the choice of Universal Turing machine to be
used as the reference machine. The problem is that there is
still subjectivity involved in this choice since what is simple
on one Turing machine may not be on another. More formally, it
can be shown that for any arbitrarily complex string $x$ as
measured against the UTM $U$ there is another UTM machine $U'$
for which $x$ has Kolmogorov complexity $1$. This result seems
to undermine the entire concept of a universal simplicity
measure but it is more of a philosophical nuisance which only
occurs in specifically designed pathological examples. The
Turing machine $U'$ would have to be absurdly biased towards
the string $x$ which would require previous knowledge of $x$.
The analogy here would be to hard-code some arbitrary long
complex number into the hardware of a computer system which is
clearly not a natural design.

To deal with this case we make the soft assumption that the
reference machine is {\em natural} in the sense that no such
specific biases exist. Unfortunately there is no rigorous
definition of {\em natural} but it is possible to argue for a
reasonable and intuitive definition in this context.
A universal Turing machine should be considered natural if it
does not contain any extreme biases. In other words if it does
not make any arbitrary, intuitively complex strings, appear
simple. It is possible to make a reasonable judgement about
this but it is preferable if there is a formal criterion which
can be applied.

One possible criterion is that a reference machine is {\em
natural }if there is a short interpreter/compiler for it on
some predetermined and universally agreed upon reference
machine. If a machine did have an inbuilt bias for any complex
strings then there could not exist a short
interpreter/compiler. If there is no bias then we assume it is
always possible to find a short compiler.

A bit more formally this is known as the short compiler
assumption \cite{Hutter:04uaibook} and can be stated as
follows. ``Given two {\em natural} Turing-equivalent formal
systems $F1$ and $F2$ there always exists a single {\em short}
 program $I$ on $F2$ that is capable of interpreting all $F1$
programs''. This assumption is important in establishing the
universality of Kolmogorov's complexity measure. If string $x$
has Kolmogorov complexity $K_{F1}(x)$ relative to system $F1$
then the upper bound of $K_{F2}(x)$ is $K_{F1}(x)+\length(I)$
where $\length(I)$ is the length of the short interpreter. This
follows simply from the fact that any $x$ can be
encoded/described on $F2$ by using the encoding for $F1$
followed by the interpreter. There may of course be shorter
descriptions but the shortest description is clearly at most
this length. Analogous reasoning shows that $K_{F1}(x)\leq
K_{F2}(x)+O(1)$. This means that the Kolmogorov complexity of a
string with respect to some system $F$ will be the same for any
{\em natural} $F$, within a reasonably small constant which is
independent of the string being measured.

To make the above criterion formal it is necessary to quantify
this concept of {\em short}. The larger it is the more flexible
this definition of {\em natural} becomes. But there is a still
a serious problem. The definition relies on the existence of
``some predetermined and universally agreed on reference
machine'' which there is currently no consensus about. In
deciding on which UTM to use for this definition it seems
reasonable to choose the `most' natural UTM but this is
obviously a circular endeavor. It may be argued
\cite{Hutter:04uaibook} that the precise choice of machine is
not of critical importance as long as it is intuitively natural
since, by the short compiler assumption, the complexity will
remain approximately equal. From this perspective the practical
and theoretical benefit of having some final fixed reference
point outweighs the importance of making this fixed reference
point `optimal' in some sense, since it has little practical
impact and appears to be philosophically unsolvable.

This issue is one of the outstanding problems in algorithmic
information theory \cite{Hutter:09aixiopen}. Fixing a reference
machine would fix the additive and multiplicative constants
that occur in many results and draw criticism to the field.
Although it by no means solves the problem there is another
useful way to view the issue.

The Kolmogorov complexity of a string depends only on the
functionality of the universal reference machine and not its
exact construction. That is, if there are two machines that,
given the same input, always produce the same output, then they
are said to be functionally equivalent and will result in the
same Kolmogorov complexity for any string. The purpose of a
universal Turing machine is only to simulate the Turing machine
that is encoded as input and therefore the output of a
universal Turing machine is uniquely defined by the Turing
machine it is simulating (and the input for this Turing
machine). This means that if two different UTM's simulate the
same Turing machine then they must produce the same output. If
they both use the same encoding scheme then simulating the same
Turing machine corresponds to having the same input and hence
they must be functionally equivalent since the same input will
always produce the same output. Since we only care about
functionality, this observation shows that the choice of
universal reference machine is equivalent to a choice of
encoding scheme. The significance of this is that it is easier
to argue for an intuitively natural encoding scheme than an
intuitively natural Turing machine.

\section{How to Choose the Prior}\label{cha:prior}

As previously shown, the Bayesian framework results in
excellent predictions given a model class $\M$ that contains
the true environment and a reasonable prior $w_\nu$ assigned to
each hypothesis $\nu\in\M$. Unfortunately the framework gives
no rigorous general method for selecting either this class or
the priors. In the Bayesianism section we briefly discussed how
to make a reasonable choice of model class and prior. Here we
examine the prior in further detail; specifically general
approaches and possible issues.

\subsection{Subjective versus Objective Priors}

A good prior should be based on reasonable and rational beliefs
about all possible hypotheses before any evidence for them has
been seen. This statement is somewhat ambiguous however since
it is debatable which observations can be considered evidence.
When looking at universal induction, every observation we make
is potentially relevant; for particular experiments it can be
hard to know in advance what is relevant. This stems
fundamentally from the subjective interpretation of probability
at the heart of Bayesianism. Because of this, a choice of prior
often belongs to one of two categories. Firstly there are
objective priors based on rational principles which should
apply to anyone without any prior relevant knowledge. Secondly
there are subjective priors that attempt to capture an agent's
personal relevant experience or knowledge. For example a
subjective prior for some experiment may be significantly
influenced by experience with similar experiments.

Solomonoff induction can deal with both approaches, leading to
a model of universal induction. Obviously we require some form
of objective prior before any observations have been made,
since there is no available information to create a subjective
prior. From this point on every observation is used to update
beliefs and these new beliefs could be interpreted as
subjective priors based on past experience, used for the next
problem.

Consider again drawing black or white balls from an urn with
some unknown ratio. Assume you start with a prior biased
towards believing the ratio is $50:50$. After observing $10$
black balls in a row initially you may interpret the situation
in two equivalent ways. Either you are $10$ balls into this
experiment and your belief has changed, or you are starting the
experiment again but now your prior is skewed to a ratio with
more black balls. More generally your posterior belief $P(H|E)$
about each hypothesis $H$ after observing $E$ becomes your new
prior $w_E(H)$ for the observations following $E$.

\subsection{Indifference Principle}

Quantifying Epicurus's principle of multiple explanations leads
to the indifference principle which assumes that if there is no
evidence favoring any particular hypothesis then we should
weight them all as equally likely. When told that an urn
contains either all black balls or all white balls and no other
information, it seems natural to assign a probability of $0.5$
to each hypothesis before any balls have been observed. This
can be extended to any finite hypothesis class by assigning
probability $\frs{1}{|\M|}$ to each hypothesis where $|\M|$ is
the number of hypotheses in $\M$.

For a continuous hypothesis class the analogous approach is to
assign a uniform prior density which must integrate to $1$ to
be a proper probability density. This means that if the
fraction of black balls in the urn is $\t\in[0,1]$ with no
extra information, we assign a uniform density of $w(\t)=1$ to
all $\t$, as seen in the derivation of the rule of succession.
This does not mean that the agent is certain in any parameter,
rather for any interval $(a,b)\subseteq[0,1]$ the belief that
$\t\in(a,b)$ is given by the integral $\int_a^b w(\t)d\t=b-a$.
Specifically, the belief in any exact value $\t$ is zero which
gives rise to the zero prior problem and hence the confirmation
problem as discussed previously.

Furthermore, in some situations this indifference principle can
not be validly applied at all. For a countably infinite class
$\M$, the probability $\frs{1}{|\M|}$ is zero which is invalid
since the sum is identically zero. Similarly, for a continuous
parameter over an infinite (non-compact) range, such as the
real numbers, the density must be assigned zero which is again
invalid since the integral would also be zero. Even when it can
be applied, two further issues that often arise are
reparametrization invariance and regrouping invariance.

\subsection{Reparametrization Invariance}

The idea of indifference and hence uniformity in a prior seems
quite straight forward but a problem occurs if a different
parametrization of a space is used. The problem is that if
there are multiple ways of parametrizing a space then applying
indifference to different choices of parametrization may lead
to a different prior.

Imagine $1000$ balls are drawn, with replacement, from an urn
containing black and white balls. The number of black balls
drawn out of these $1000$ samples must obviously lie somewhere
between $0$ and $1001$ and we denote this number $k$. It may be
argued that since we know nothing it is reasonable to assume
indifference over the $1000$ possible values of $k$. This would
result in a roughly equal prior over all values of $\t$. More
specifically it would assign a prior of $\frs{1}{n+1}$ to every
value in the set $\{\t=\frs{i}{n}|i\in\{0,...,n\}\}$, where $n$
is the total number of trials, in this case $1000$.

On the other hand it may be argued that it is equally plausible
to have a prior that is indifferent over every observable
sequence of $1000$ black or white balls. However, if each of
these $2^{1000}$ possible sequences are assigned equal
probability then a very different prior is created. The reason
for this is that there are far more possible sequences which
contain approximately $500$ black balls than there are
sequences that contain nearly all black balls or nearly no
white balls. In the extremes there is only one possible
sequence for which $k=0$ and one possible sequence for which
$k=1000$, but there are ${1000 \choose
500}=\frac{1000!}{500!^2}\approx 10^{300}$ possible sequences
for which $k=500$. This means that indifference over every
possible sequence leads to a prior over $\t$ that is strongly
peaked around $\t=0.5$. This peak is sharper for a higher
number of trials.

Similarly it may seem equally valid to assume indifference over
either $\t$ or $\sqrt{\t}$ which would again lead to
different priors. In some situations the `correct' choice of
parametrization may be clear but this is certainly not always
the case. Some other principles do not have this issue.
Principles that lead to the same prior regardless of the
choice of parametrization are said to satisfy the
reparametrization invariance principle (RIP). Formally the
criterion for the RIP is as follows.

By applying some general principle to a parameter $\t$ of
hypothesis class $\M$ we arrive at prior $w(\t)$. For example
$\t\in[0,1]$ as above leads to $w(\t)=1$ by indifference. We
now consider some new parametrization $\t'$ which we assume is
related to $\t$ via some bijection $f$. In this case we
consider $\t'=f(\t)=\sqrt{\t}$. Now there are two ways to
arrive at a prior which focuses on this new parameter $\t'$.
Firstly we can directly apply the same principle to this new
parametrization to get prior $w'(\t')$. For the indifference
principle this becomes $w'(\t')=1$. The second way is to
transform the original prior using this same bijection. When
the prior is a density this transformation is formally given by
$\tilde w(\t')=w(f^{-1}(\t'))df^{-1}(\t')/d\t'$. For
$\t=f^{-1}(\t')=\t'^2$ this leads to $\tilde
w(\t')=2\sqrt{\t}$. If $w'=\tilde w$ then the principle
satisfies the reparametrization invariance principle. It is
clear from the example of $\t'=\sqrt{\t}$ that the indifference
principle does not satisfy RIP in the case of densities.

It does however satisfy RIP for finite model classes $\M$. This
is because for a finite class the prior $w_\nu=\frs{1}{|\M|}$
for all $\nu\in\M$ and hence $w(\nu)=\frs{1}{|\M|}$ for any
reparametrization $f$. A reparametrization in a finite class is
essentially just a renaming that has no affect on the
indifference.

\subsection{Regrouping Invariance}

Regrouping invariance can be thought of as a generalization of
the concept of reparametrization invariance. This is because
reparametrization involves a function that is a bijection and
hence every instance of the transformed parameter corresponds
to one and only one instance of the original parameter.
Regrouping on the other hand involves a function that is not
necessarily bijective and hence can lead to a many to one or
one to many correspondence.

For example, the non-bijective function $\t'=f(\t)=\t^2$ for
$\t\in[-1,1]$ leads to the regrouping $\{+\t,-\t\}\leadsto
\{\t^2\}$. A more intuitive example can be seen if we again
consider the observation of ravens. Previously only the binary
information of a black or a non-black raven was recorded but we
might also be interested in whether the raven was black, white
or colored. In this case the population can not be parametrized
with only one parameter $\t$ as before. For i.i.d.\ data in
general there needs to be as many parameters (minus one
constraint) as there are possible observations. These
parameters specify the percentage of the population that is
made up by each observation. For the binary case only the
percentage of black ravens $\t$ was used but since the
parameters must sum to one the percentage of non-black ravens
is implicity defined by $(1-\t)$. Formally, for an i.i.d.\
space with $d$ possible observations the parameter space is
$\triangle^{d-1}:=\{\vec\t\equiv(\t_1,...,\t_d)\in[0,1]^d:\sum_{i=1}^d\t_i=1\}$.
In the binary case, the probability of a string $x_{1:n}$ with
$s$ successes and $f$ failures was given by
$P(x_{1:n}|\t)=\t^s(1-\t)^f$. In the case of $d$ observations
the probability of $x_{1:n}$, with $n_{i}$ occurrences of
observation $i$, is analogously given by
$P(x_{1:n}|\vec\t)=\prod_{i=1}^d\t_{i}^{n_{i}}$.

The regrouping problem arises when we want to make inferences
about the hypothesis ``all ravens are black'' when the setup is
now to record the extra information of whether a raven is
colored or white. Intuitively recording this extra information
should not affect the outcome of reasoning about black ravens
but unfortunately it does if we apply the principle of
indifference. When we make an inference that only looks at the
`blackness' of a raven, the observations are collapsed into
blackness or non-blackness as before by mapping black to
success and either white or colored to failure. We can then use
the binary framework as before with
$P(x_{1:n}|\t)=\t^s(1-\t)^f$. However, since we assumed
indifference over the parameter vectors in $\triangle^2$ this
regrouping means that the prior belief is skewed towards higher
proportions of non-black ravens and is therefore no longer
indifferent. Indeed, $w(\vec\t)=${\em constant} for
$\vec\t\in\triangle^2$ leads to $\tilde w(\t')=2(1-\t')\neq
1=w'(\t')$ for $\t'\in[0,1]$. This means that the indifference
principle is not invariant under regrouping.

Because the function $f$ is not bijective anymore, the
transformation of the prior $w(\t)$ to some new parametrization
$\t'$ now involves an integral or sum of the priors over all
value of $\t$ for which $f(\t)=\t'$. Formally, for discrete
class $\M$ we have $\tilde w_{\t'}=\sum_{\t:f(\t)=\t'}w_\t$,
and similarly for continuous parametric classes we have $\tilde
w(\t')=\int\delta(f(\t)-\t')w(\t)d\t$. As with
reparametrization invariance before, for a principle to be
regrouping invariant, we require that $\tilde w(\t')=w'(\t')$
where $w'(\t')$ is obtained by applying the same principle to
the new parametrization.

It is generally considered highly desirable that a principle
for creating priors is regrouping invariant but intuitively it
seems that this invariance is a difficult property to satisfy.
Attempting to satisfy it for all possible regrouping leads to a
predictor that is unacceptably overconfident. Formally this
means that $\xi(1|1^n)=1$ which, as Empiricus argued, is an
illogical belief unless we have observed every instance of a
type. In fact, it was shown \cite{Wallace:05} that there is no
acceptable prior density that solves this problem universally.

Luckily the universal prior is not a density and one can show
it approximately satisfies both, reparametrization and
regrouping invariance \cite{Hutter:07uspx}.

\subsection{Universal Prior}\label{sec:uprior}

The universal prior is designed to do justice to both Occam and
Epicurus as well as be applicable to any computable
environment. To do justice to Epicurus' principle of multiple
explanations we must regard all environments as possible, which
means the prior for each environment must be non zero. To do
justice to Occam we must regard simpler hypotheses as more
plausible than complex ones. To be a valid prior it must also
sum to (less than or equal to) one. Since the prefix Kolmogorov
complexity satisfies Kraft's inequality, the following is a
valid prior.
\[
  w_\nu^U \;:=\; 2^{-K(\nu)}
\]
This prior is monotonically decreasing in the complexity of $\nu$ and
is non-zero for all computable $\nu$.

This elegant unification of the seemingly opposed philosophies
of Occam and Epicurus is based only on these universal
principles and the effective quantification of simplicity by
Kolmogorov. The result is a prior that is both intuitively
satisfying and completely objective.

When the bounds for Bayesian prediction of
Subsection~\ref{sec:conv} are re-examined in the context of the
Universal Prior we see that the upper bounds on the deviation of
the Bayesian mixture from the true environment are
$\ln(w_\mu^{-1})=\ln(2^{K(\mu)})=K(\mu)\ln(2)$. This means it
is proportional to the complexity of the true environment which
is not surprising. In simple environments the convergence is
quick while in complex environments, although the framework
still performs well, it is more difficult to learn the exact
structure and hence convergence is slower.

The universality of Kolmogorov complexity bestows the universal
prior $w_\nu^U$ with remarkable properties. First, any other
reasonable prior $w_\nu$ gives approximately the same or
weaker bounds. Second, the universal prior approximately
satisfies both, reparametrization and regrouping invariance
\cite{Hutter:07uspx}. This is possible, since it is not a
density.

\section{Solomonoff Universal Prediction}\label{cha:solomonoff}

Ray Solomonoff, born July 25th 1926, founded the field of
algorithmic information theory. He was the first person to
realize the importance of probability and information theory in
artificial intelligence. Solomonoff's induction scheme
completes the general Bayesian framework by choosing the model
class $\M$ to be the class of all computable measures and
taking the universal prior over this class. This system not
only performs excellently as a predictor, it also conforms to
our intuition about prediction and the concept of induction.

It should be appreciated that according to the Church-Turing
thesis, the class of all computable measures includes
essentially any conceivable natural environment. John von
Neumann once stated {\em ``If you will tell me precisely what
it is that a machine cannot do, then I can always make a
machine which will do just that''}. This is because, given any
``precise'' description of a task we can design an algorithm to
complete this task and therefore the task is computable.
Although slightly tongue in cheek and not quite mathematically
correct, this statement nicely captures the universality of the
concept of computability. There are of course imprecise ideas
such as love or consciousness which can be debated, but if a
system consists of clear-cut rules and properties then it is
usually computable. According to the laws of physics our world
is governed by precise, although not necessarily deterministic,
rules and properties and indeed any physical system we know of
is computable.

Unfortunately this is actually a slight simplification and the
concept of infinity causes some technical issues. This is
because, for a task to be formally computable it is required
not only that it can be described precisely but also there is a
program that completes this task in finite time, in other
words it will always terminate. Due to the fundamentally finite
nature of our universe this is not usually an issue in physical
systems. However, in a more abstract environment such as the
platonic world of math, the existence of infinite sets,
infinite strings and numbers with infinite expansions means
that this termination becomes more of an issue. In particular
the sum over an infinite set of environments present in Bayes
mixture can be incomputable and therefore one has to introduce
a slightly broader concept of semi-computable. For a task to be
semi-computable there must exist a program that will
monotonically converge to the correct output but may never
terminate. More formally, there is a Turing machine that
outputs a sequence of increasing values which will eventually
be arbitrarily close to the correct value, however we don't know
how close it is and it may never output the correct value
and/or may never halt.

\subsection{Universal Bayes Mixture}

Due to technical reasons it is convenient to choose $\M$ to be
the class of all semi-computable so-called semi-measures. This
extension of the class does not weaken its universality since
any computable environment is also semi-computable. In fact we
could further extend it to include all non-computable
environments without changing the result because the universal
prior for any non-computable environment is zero and therefore
the prediction from any non-computable environment does not
contribute. From here on this universal class is denoted $\M_U$
and the Bayesian mixture over this class using the universal
prior, called the universal Bayesian mixture, is denoted
$\xi_U$. Formally this is defined as before
\[
  \xi_U(x) \;=\; \sum_{\nu\in\M_U}w_\nu^U\nu(x)
\]
where $w_\nu^U=2^{-K(\nu)}$ is the universal prior. Since the
class $\M_U$ is infinite, the Bayesian mixture $\xi_U$ contains
an infinite sum and therefore it is not finitely computable. It
can however be approximated from below which means it is a
semi-computable semi-measure and therefore a member of $\M_U$
itself. This property is one of the reasons that the extended
model class was chosen. The proof that $\xi_U$ is
semi-computable is non-trivial and is important in establishing
its equivalence with the alternative representation $M$
below.

\subsection{Deterministic Representation}\label{sec:det}

The above definition is a mixture over all semi-computable
stochastic environments using the universal prior as weights.
It is however possible to think about $\xi_U$ in a completely
different way. To do this we assume in this subsection that the
world is governed by some deterministic computable process.
Since it is computable this process can be described by some
program $p$ which is described using less than or equal to
$\ell$ bits. This is possible since every program must have
finite length, however $\ell$ may be arbitrarily large. This
upper bound $\ell$ on the length of $p$ must of course be in
relation to some universal reference Turing machine $U$, since
each UTM uses a different encoding scheme which may affect the
length of $p$.

As before the aim is to make the best possible predictions
about observations $x$. Again this is done using a universal
distribution over all binary strings $x$ which reflect our
beliefs in seeing those strings, even though the true
environment $p$ produces only one predetermined output string
under this interpretation. In order to make the distribution
universal it is important that there is no bias or assumptions
made about the structure of the world. The string $x$
represents the current observations or equivalently the initial
output string of the true environment. At any point however we
can not know whether the program has halted and the output
string is complete or whether there is more output to come.
More generally we say that a program $p$ produces $x$ if it
produces any string starting with $x$ when run on $U$. Formally
this is written as $U(p)=x*$. For this purpose programs that
never halt are also permitted. Once it has produced a string
starting with $x$ it may continue to produce output
indefinitely.

The probability of observing $x$ can be computed using the rule
$P(x)=\sum_p P(x|p)P(p)$. This sum is over all programs of
length $\ell$. To remain unbiased, Epicurus's principle is
invoked to assign equal prior probability to each of these
environments a priori. Since there are $2^{\ell}$ of these
programs they each get assigned probability
$(\frac{1}{2})^\ell$, so $P(p)=2^{-\ell}$ for all $p$ of length
$\ell$. As each of these programs $p$ is deterministic the
probability $P(x|p)$ of producing $x$ is simply $1$ if it does
produces $x$ or $0$ if it doesn't. Because $P(x|p)=0$ for any
$p$ where $U(p)\neq x*$ these programs can simply be dropped
from the sum in the expression for $P(x)$ so it now becomes
\[
  P(x) \;=\; \sum_{p:U(p)=x*}2^{-\ell}
\]
This sum is still only over programs of length $\ell$, but
there may be many shorter programs which also produce $x$.
Since the assumption was only that the world is governed by
some program with less than or equal to $\ell$ bits there is no
reason not to also consider these shorter programs. Fortunately
these programs are automatically accounted for due to the
technical setup of the definition. If there is a program $p$
with length less than $\ell$ then we can simply pad it out
until it has length $\ell$ without affecting how it operates
and hence what it outputs. This padding can consist of any
arbitrary binary string and therefore, importantly, the shorter
$p$ is the more ways there are to pad it out to the full
length. To be precise, for any program $p$ with
$\length(p)\leq \ell$ there are exactly $2^{\ell-\length(p)}$
different ways to extend it and each of these
$2^{\ell-\length(p)}$ programs now have length $\ell$ and
output $x$. This means that any program $p$ of length
$\length(p)$ contributes
$2^{\ell-\length(p)}\times2^{-\ell}=2^{-\length(p)}$ to the
above sum. This property means that the sum is actually over
any program with length less than or equal to $\ell$ and the
contribution of a program depends on its true length.

Now since $\ell$ can be arbitrarily large we extend this sum to
be over any program of any length. Also, in order to avoid
counting the same `core' programs multiple times we introduce
the concept of a minimal program $p$ that outputs a string
$x$. A program is minimal if removing any bits from the end
will cause it to not output $x$, and hence programs with
arbitrary padding are clearly not minimal. We can therefore
rewrite the above sum as
\[
  M(x) \;:=\; \sum_{p:U(p)=x*}2^{-\length(p)}
\]
where the sum is now over all minimal programs $p$ of arbitrary
length. We call $M(x)$ the universal probability of $x$. $M(x)$
can also be seen as the frequentist probability of $x$ as it
corresponds to the total number of programs (minimal or
non-minimal) of length $\ell$ that produce $x$, divided by the
total number of programs of length $\ell$, in the limit as
$\ell\to\infty$.

This is a highly technical explanation of $M(x)$ but there is a
simpler way to think about it. The set of these arbitrary
programs, without the restriction that they must produce $x$,
is actually the set of all possible binary strings. This is
because any binary input string for a universal Turing machine
is considered to be a program, even though the vast majority of
these programs will not produce any useful or meaningful
result. The contribution of any particular program $p$ with
length $\length(p)$ is $2^{-\length(p)}$. It is no coincidence
that this corresponds to the probability of producing this
program by simply flipping a coin for each bit and writing a
$1$ for heads and a $0$ for tails. This means that the
contribution of a particular program corresponds to the
probability that this program will appear on the input tape
when the Universal Turing machine is provided with completely
random noise. Now the set of programs that produce $x$ is the
set of programs which contribute to $M(x)$. Therefore $M(x)$ is
actually the probability of producing $x$ when random noise is
piped through some universal reference Turing machine $U$. This
alternative representation of $M(x)$ leads to some interesting
insights.

Firstly, this entirely different approach to finding a
universal probability of $x$ turns out to be equivalent to
$\xi_U$. Even though $M(x)$ considers only deterministic
environments and $\xi_U(x)$ sums over all semi-computable
stochastic environments, they actually coincide,
within a multiplicative constant which can be eliminated and
henceforth will be ignored \cite{Hutter:04uaibook}. The
intuitive reason for this is that, roughly speaking, the
stochastic environments lie in the convex hull of the
deterministic ones. To see this consider the two points on the
real line between $1$ and $2$. The convex hull of these two
points is the set of points $c$ such that $c=a\cdot1+b\cdot2$
where $a+b=1$ and $a,b\geq 0$. In other words any point in the
convex hull is a mixture of $1$ and $2$ where the mixing
coefficients, like a probability distribution, must sum to $1$.
This set of points is simply the interval $[1,2]$. Given three
points in a plane, the convex hull is the filled-in triangle
with vertices at these points. In the more abstract setting of
deterministic environments the same intuition holds. As a
simple example imagine you must open one of two doors and you
know for a fact that one door has a cat behind it and the other
door has a dog behind it. You are unsure which door holds which
animal however, but you have a $60\%$ belief that the first
door is the door with the cat. This means that under your own
Bayesian mixture estimation your belief that the first door
holds a cat is $0.6$. This may seem trivial but it illustrates
that by mixing deterministic environments, such as the two
possibilities for the doors, the resulting estimation may be
stochastic. The same principle holds for all computable
deterministic environments. Every stochastic environment is
equivalent to some mixture of deterministic environments. Also,
since any mixture of stochastic environments is itself a
stochastic environment, it is also equivalent to some mixture
of deterministic environments.

The equivalence of $\xi_U$ and $M$ is particularly surprising
considering that $\xi_U(x)$ uses the universal prior, which
favors simple environments, while $M(x)$ is based on Epicurus's
principle which is indifferent. On closer examination however,
there is an interesting connection. In $M(x)$ the shortest
program $p$ which produces $x$ clearly has the greatest
contribution to the sum, since $2^{-\length(p)}$ is maximal
when $\length(p)$ is minimal. In fact another program $q$ which
has length only $1$ bit greater contributes only half as much
since $2^{-(\length(p)+1)}=\frac{1}{2}\times 2^{-\length(p)}.$
In general a program with only $n$ more bits contributes only
$2^{-n}$ as much as the shortest. It turns out that the
contribution from this shortest program actually dominates the
sum. But the length of the shortest program $p$ that produces
$x$ is the Kolmogorov complexity $K(x)$ and therefore $M(x)$ is
approximately equal to $2^{-K(x)}$ which is the universal prior
for $x$. It is interesting that by starting with the
indifference principle and using only logical deduction in the
framework of Turing machines we arrive at a predictor that
favors simple descriptions in essentially the same way as
advocated by Occam's razor. This can be seen as both an
intuitively appealing characteristic of $M(x)$ and as a
fundamental justification for Occam's razor
\cite{Hutter:10ctoex}.

\subsection{Old Evidence and New Hypotheses}

One of the problems with the Bayesian framework is dealing with
new hypotheses $H$ that were not in the original class $\M$.
In science it is natural to come up with a new explanation of
some data which cannot be satisfactorily explained by any of
the current models. Unfortunately the Bayesian framework
describes how to update our belief in a hypothesis according to
evidence but not how to assign a belief if the hypothesis was
created to fit the data. Several questions come to mind in
trying to solve this problem such as ``should old evidence be
allowed to confirm a new hypothesis?'' or ``should we try to
reason about our belief as if the evidence did not exist?''.

By choosing the universal class $\M_U$ this problem is formally
solved. Theoretically it can no longer occur since this class
is complete in the sense that it already contains any
reasonable hypothesis. Since the full mixture is however
incomputable it is likely that an approximation will only
consider a subset of this class, but this is not a problem as
the universal prior is unaffected. If a `new' hypothesis $H$ is
considered then it is simply assigned its universal prior
$2^{-K(H)}$, and the evidence can then be used to update this
prior as if it had been in the original approximation class.
Although hypotheses can still be constructed to fit data they
will automatically be penalized if they do so in a complex way.
This is because if $H$ is naively used to account for $E$ then
$K(H)\geq K(E)$ and hence, if $E$ is complex, $2^{-K(H)}$ will be
small. A hypothesis that naively accounts for $E$ is similar
to a program that `hard codes' the string corresponding to
$E$.

For example in the ancient Greek's geocentric model of
planetary motion it was discovered that using perfect spheres
for the orbits of the planets did not fit with the observed
data. In order to maintain the highly regarded perfect sphere
and the ideological point of the earth being the center of the
universe, complex epicycles were added to the model to fit all
observations. Finally, in the year of his death in 1543,
Nicolaus Copernicus published a far more elegant model which
sacrifices perfect spheres and placed the sun at the center of
the solar system. Despite resistance from the church the
undeniable elegance of this solution prevailed.

Generally, by following this method, a new hypothesis that
accounts for some data in an elegant and general manner will be
believed far more than a new hypothesis that is overly biased
towards this particular data. This seems reasonable and
corresponds with common practice.

\subsection{Black Ravens Paradox Using Solomonoff}\label{sec:BR}

We will now look at the black raven's paradox in relation to
Solomonoff induction. First we will show that Solomonoff
induction is able to confirm the hypothesis ``all ravens are
black''. We will then comment on the relative degree of
confirmation under Solomonoff.

Laplace looked at the binary i.i.d.\ case of drawing balls from
an urn. The confirmation problem in this case arose from the
zero prior in any complete hypothesis due to the use of a prior
density over the continuous class of hypotheses
$\M=\{\t|\t\in[0,1]\}$. Confirmation for the hypothesis $\t=1$
was made possible by assigning an initially non-zero point
probability mass to $w(\t)$ for $\t=1$. In this case there are
only two possible observations, a white ball or a black ball.
The parameter $\t$ represents the proportion of black balls but
we also have another parameter representing the proportion of
white balls. This second parameter is implicity given by
$(1-\t)$ due to the constraint that they must sum to one. The
two constraints on these parameters, that they must sum to one
and lie in the interval $[0,1]$, can be thought of as a
one-dimensional finite hyperplane which is equivalent to a unit
line.

In the black ravens paradox we have to define two predicates,
Blackness and Ravenness. There are four possible
observations: These are a black raven $BR$, a non-black raven
$\overline{B}R$, a black non-raven $B\overline{R}$ and finally
a non-black non-raven $\overline{BR}$, i.e.\ the observation
alphabet is
$\X=\{BR,\overline{B}R,B\overline{R},\overline{BR}\}$. Each of
these types has an associated parameter that represents the
proportion of the entire population which belong to each of the
respective types. These parameters are denoted $\vec\t \equiv
(\t_{BR},\t_{\overline{B}R},\t_{B\overline{R}},\t_{\overline{BR}})$
respectively. This makes the setup significantly more complex
to work with. Since any object must belong to one and only one
of these types the corresponding parameter values must sum to
one. A complete hypothesis is given by any valid assignment of
each of these four parameters and therefore the model class is
given by the following:
\[
  \M_{\triangle^3} \;:=\; \{\vec\t\in[0,1]^4 \,:\, \t_{BR}+\t_{\overline{B}R}+\t_{B\overline{R}}+\t_{\overline{BR}}=1\}
\]

\begin{wrapfigure}{r}{0.3\textwidth}
  \includegraphics[width=0.3\textwidth]{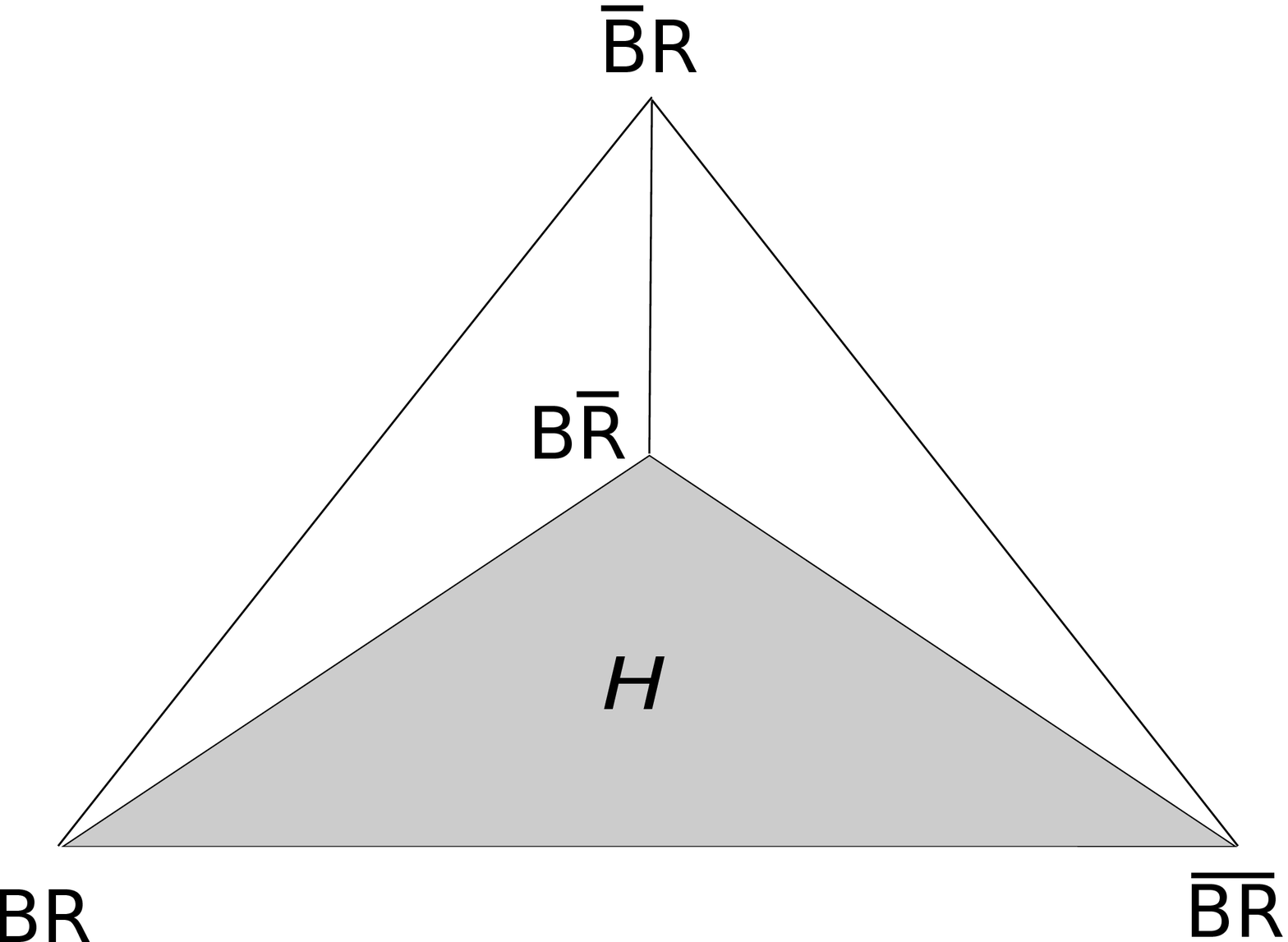}
  \vspace{-5ex}
\end{wrapfigure}
This space can be visualized as a three-dimensional finite
hyperplane which is equivalent to a 3-simplex, as in
the figure on the right, where each
vertex corresponds to the hypothesis that one of the parameters
make up all observations. Vertex $BR$ corresponds to the
hypothesis that all objects are black ravens, Vertex
$\overline{BR}$ corresponds to the hypothesis that all objects
are non-black non-ravens and so forth. The point in the very
center of the probability simplex corresponds to the hypothesis
$\t_{BR}=\t_{\overline{B}R}=\t_{B\overline{R}}=\t_{\overline{BR}}=\frs{1}{4}$.
Since non-black non-ravens make up the vast majority of objects
in the real world, the correct hypothesis corresponds to a
point very close to the vertex given by $\t_{\overline{BR}}=1$.

We are now interested in the {\em partial} hypothesis $H_b$
given by the statement ``all ravens are black''. This
represents the set of any points where the statement is true.
But the statement ``all ravens are black'' is logically
equivalent to the statement ``there are no non-black ravens'',
and this occurs exactly when $\t_{\overline{B}R}=0$. Therefore
our partial hypothesis represents the set
$H_b=\{\vec\t\in\M_{\triangle^3} \,:\, \t_{\overline{B}R}=0\}$.
In the figure above this corresponds to the entire face $H$ of
the simplex that lies opposite the vertex at $\overline{B}R$.

The question remains whether we are able to confirm this
hypothesis using the universal prior. This requires that the
hypothesis $H_b$ is assigned a non-zero prior. The universal
prior in this case depends on the complexity of the values of
each of the parameters. Intuitively, the integers $0$ and $1$
are the simplest numbers in the interval $[0,1]$. Therefore the
universal prior $P(\vec\t) = w_{\vec\t}^U = 2^{-K(\vec\t)}$ is
going to favor hypothesis that have parameter values of $0$ or
$1$. The hypotheses corresponding to each vertex have one
parameter set to one and the rest set to zero, e.g.\
$(\t_{BR},\,\t_{\overline{B}R},\,\t_{B\overline{R}},\,\t_{\overline{BR}})=(1,0,0,0)$.
These are the simplest valid assignments of the parameters and
therefore have the highest prior. If a hypothesis lies on the
face of the simplex then it implies that one of the parameters
must be zero, as in $H_b$. Therefore hypotheses lying on a face
are also implicitly favored. Specifically, there are hypotheses
corresponding to points on a face that are assigned a non-zero
prior and therefore, $w_b = P(H_b) = \;\;\int\!\!\!\!\!\!\sum_{h\in
H_b}P(h) = \sum_{\vec\t\in H_b} 2^{-K(\vec\t)} > 0$. This means
$H_b$ is assigned a non-zero prior as required. In fact any
hypothesis that has computable values assigned to its
parameters has a non-zero prior. Since these `computable'
hypotheses are dense in the simplex, essentially any reasonable
subset of this model class is assigned a non-zero prior.

The choice of class $\M_{\triangle^3}$ implicitly assumes that
observations are sampled i.i.d: The probability of observing a
sequence of $n$ (non)black (non)ravens objects
$x_{1:n}\in\X^n$ if $\vec\t$ is the true parameter is
$P(x_{1:n}|\vec\t) = \prod_{x\in\X}\t_x^{n_x}$, where $n_x$ is
the number of times, $x$ has been observed. This i.i.d.\
assumption will be relaxed below.

The physical hypothesis $H_b$ refers to the unobservable parameters
$\vec\t$. Similarly as in the binary confirmation problem in
Section~\ref{sec:confirm}, we can introduce an equivalent
observational hypothesis $H'_b := \{x_{1:\infty}:\forall
t\;x_t\neq\overline{B}R\} \equiv
\{BR,B\overline{R},\overline{BR}\}^\infty$ that consists of
all possible observation sequences consistent with the
hypothesis that all ravens are black. One can show that
$H_b$ as well as $H'_b$ asymptotically get fully confirmed, i.e.\
\[
  P(H_b|x_{1:n}) \;=\; P(H'_b|x_{1:n}) \;\stackrel{a.s.}{\longrightarrow}\; 1
  \qmbox{if $x_{1:\infty}$ is sampled from any $\vec\t\in H_b$}
\]

Let us now consider the universal mixture $\xi_U=M$, which is
not limited to i.i.d.\ models. Clearly $\M_U$ includes all
i.i.d.\ environments with computable parameter $\vec\t$, hence
$M$ should be able to confirm the black raven hypothesis too.
Indeed, consider any computable probability distribution $\mu$
consistent with $H'_b$, i.e.\ $\mu(x)=0$ for $x\not\in
\{BR,B\overline{R},\overline{BR}\}^*$.
Let $x_{1:\infty}$ be a sequence sampled from $\mu$, which by
definition of $\mu$ does not contain any non-black ravens. Then
the convergence in total variation result from
Subsection~\ref{sec:totbounds} (choose $A=H'_b$ and
exploit $\mu(H'_b|x_{1:n})=1$) implies
\[
  M(H'_b|x_{1:n}) \;\longrightarrow\; 1
  \qmbox{with $\mu$ probability 1}
\]
That is, $M$ strongly confirms the black raven hypothesis under any
reasonable sampling process, of course provided no non-black
ravens are observed. No i.i.d.\ or stationarity or ergodicity or
other assumption is necessary. This is very favorable, since
real-life experience is clearly not i.i.d.
What remains to be seen is whether $M$ also gets the absolute
and relative degree of confirmation right. But the fact that
Solomonoff induction can confirm the black raven hypothesis
under very weak sampling assumptions even in infinite populations
containing ravens and other objects is already quite unique and
remarkable.

\section{Prediction Bounds}\label{cha:bounds}

Since Solomonoff;s approach is simply the Bayesian framework
with the universal model class and prior, the bounds for Bayes
mixture from Section~\ref{cha:bayes} remain valid for $\xi_U$,
although it is now possible to say a little more. Firstly,
since the bounds assume that $\mu\in\M$ we can now say they
hold for any computable environment $\mu$. Also, since $M$ and
$\xi_U$ coincide, the same bounds for $\xi_U$ also hold for $M$.
This means that $\xi_U$ and $M$ are excellent predictors given
only the assumption that the sequence being predicted is drawn
from some computable probability distribution which, as
discussed, is a very weak assumption, and even this can be
relaxed further.

\subsection{Total Bounds}\label{sec:totbounds}

In particular the deterministic total bound from the
Bayesianism section holds with $\xi$ replaced by $M$. This
means that in the case that the true distribution $\mu$ is
deterministic the following bound holds.
\[
  \sum_{t=1}^\infty|1-M(x_t|x_{<t})| \;\leq\; \ln(w_\mu^{-1})
  \;=\; \Km(x_{1:\infty})\ln 2
\]
Since an infinite sum of positive numbers can only be finite if
they tend to zero, this means that $M(x_t|x_{<t})\to 1$. In
other words, the predictions converge to the correct
predictions. Indeed, convergence is rapid in the sense of
$1-M(x_t|x_{<t})<\frs1t$ for all but a vanishing fraction of
$t$, which implies fast convergence in practice as long as the
sequence is not too complex. The property of being able to
predict the next observation of any deterministic computable
sequence with high probability after viewing a relatively short
initial segment of this sequence is a strong result.

Solomonoff's stronger result \cite{Solomonoff:78} generalizes
this to sequences $x$ sampled from any arbitrary unknown computable
distribution $\mu$. In particular the following bound holds
\[
  \sum_{t=1}^\infty\sum_{x_{1:t}\in\mathbb{B}^t}\mu(x_{<t})\Big(M(x_t|x_{<t})-\mu(x_t|x_{<t})\Big)^2
  \;<\; K(\mu)\ln2 + O(1) \;<\; \infty
\]
which implies that if $x$ is sampled from any $\mu$, the
universal predictor $M$ rapidly converges to the true
computable environment $\mu$ with $\mu$-probability $1$.

Both results consider one-step lookahead prediction but are
easily extendible to multi-step lookahead prediction
\cite{Hutter:04uaibook}. Exploiting absolute continuity of $\mu$
w.r.t.\ $M$, asymptotic convergence can be shown even for infinite
lookahead \cite{Blackwell:62} and any computable $\mu$:
\[
  \sup_{A\subseteq\X^\infty} \big|M(A|x_{<t})-\mu(A|x_{<t})\big| \;\longrightarrow\; 0
  \qmbox{with $\mu$-probability 1}
\]

All three results illustrate that eventually $M$ will recognize
any structure present in a sequence. This sequence may be as
simple as an infinite string of $1$'s or as complex as every
$74$th digit in the binary expansion of $\pi$ which even a
human would find very difficult to recognize. In the context of
observing ravens, or days when the sun rises, the observation
string can be thought of as a string of $1$'s where $1$
represents a black raven, or sun rise, and $0$ represents a
white raven, or no sun rise. In this case the convergence
$M(1|1^n)\to1$ is particularly fast because $\Km(1^\infty)$ is
very small. Again this is not surprising: the more obvious or
simple the structure in a sequence the better $M$ performs.

\subsection{Instantaneous Bounds}

The previous bounds give excellent guarantees over some initial
$n$ predictions but say nothing about the $n$th prediction
itself. In classes of environments that are independent and
identically distributed (i.i.d.) it is also possible to prove
good instantaneous bounds but in the general case this is more
difficult. In many cases it is reasonable to think of the data
as i.i.d.\ even if it is not strictly true. For example when
observing ravens there are many factors that affect which
raven gets observed, at the very least your location, however
thinking of the observations as being drawn randomly from the
entire population is a reasonable approximation. Sometimes it
is useful to interpret uncertain or minor factors as noise over
otherwise i.i.d.\ data. Unfortunately there are also many cases
where it is not at all reasonable to consider the data as
i.i.d.\ and for these more general cases the following bound
for computable $x$ was proven in \cite{Hutter:07uspx}:
\[
  2^{-K(n)} \;\leq\; (1-M(x_n|x_{<n})) \;\leq\; C\cdot 2^{-K(n)}
\]
where $C$ is a constant which depends on the complexity of the
environment but not on $n$.

This means that the probability of an incorrect prediction
depends on how much of the sequence has been seen as well as
the complexity of the sequence. For simple strings this
probability will quickly become small but for complex strings
only a very weak guarantee can be given. If the string is very
simple, in the sense that $\Km(x_{1:\infty})=O(1)$, then $C$
will be small and hence the lower and upper bound will coincide
within an irrelevant multiplicative constant, denoted by
$\stackrel\times=$ below. In particular for
$x_{1:\infty}=1^\infty$ we get
\[
  M(0|1^n) \;\stackrel\times=\; 2^{-K(n)}
\]
Due to the non-monotonic nature of $K(n)$ this results in an
interesting property. Because $K(n)$ dips when $n$ is simple,
$M(0|1^n)$ spikes up. In other words $M$ expects a counter
example to a universal generalization after a simple number of
observations more than after a comparatively random number of
observations. Although this may seem absurd there is a certain
validity to being ``cautious'' of simple numbers.

When we examine the world around us there is structure
everywhere. Countless patterns and regularities in what we see
and hear, even if most of it evades our conscious attention.
This may seem like a vague and subjective claim but it is
actually verifiable. This underlying structure is the reason
why fractal compressors, which operate by extracting simple
recursive regularities from images, work so successfully
\cite{Barnsley:93}. Interestingly they work particularly well
on images of natural scenery because they contain large amounts
of fractal structure. In fact there is a lot of structure in
the sequences governing many natural processes such as the
growth of a human embryo which doubles with each generation, or
the arrangement of petals and leaves on certain plants which
often obey simple mathematical rules such as the Fibonacci
sequence. Not surprisingly this underlying structure is also
found in the technological world. It has been verified, using
Google to measure frequencies of integers \cite{Cilibrasi:06},
that numbers with low Kolmogorov complexity occur with high
probability on the internet.

Structure is closely related to Kolmogorov complexity. If a
process or object is highly structured then this means we can
write a short program to predict it or produce it. If a process
has very little structure then it is closer to a random string
and requires a much more verbose description or encoding. In
the context of binary prediction with error probability bounded
by $C\cdot 2^{-K(n)}$, as above, this structure is reflected by
the simplicity of $n$. The abundance of structure in the world
means that simple numbers are more likely to be significant,
which corresponds with the non-monotonic nature of $K(n)$. It
should also be noted that although the above example considers
binary prediction, analogous results hold for more general
alphabets.

The non-monotonic bound of $2^{-K(n)}$ may still seem strange
but it is important to realize that even at simple numbers our
belief does not suddenly switch from a low belief in one
observation to a high belief. Even for large simple $n$,
$2^{-K(n)}$ is still quite small, so our belief in seeing a
zero after $1'000'000$ ones is still almost negligible, but
higher than after $982'093$ ones, a relatively random number of
the same magnitude. In any case, the total bound
$\sum_{n=0}^\infty M(0|1^n)\leq O(1)$ ensures that the size and
frequency of these `peaks' are quite limited.

It should also be noted that we have made the implicit
assumption that we know exactly how many observations we have
made so far. In reality however we often don't know $n$
precisely. Consider all the black ravens you have ever
observed. It's likely that you know you have observed many
black ravens but not likely you could recall the exact number.
This means that if you were to apply the above method strictly
you would have to formalize some prior distribution about your
belief in the exact number of ravens you have observed and then
take the average of the above method according to this prior.
This averaging would smooth out the above mentioned peaks, so
our belief in observing a different object would decrease more
uniformly. In many real world situations where the above bound
seems absurd, uncertainty in our precise observation history
would cause a more uniform decrease as these peaks are washed
out by Bayesian averaging.

\subsection{Future Bounds}

When looking at an agent's performance it is often important to
consider not only the total and instantaneous performance but
also the total future performance bounds. In other words it can
be important to estimate how many errors it is going to make
from now on. Consider a general agent learning how to play
chess from scratch. Obviously it will make a large number of
errors initially as it does not even know the rules of the game
so most of its moves will be invalid. More interesting is to
examine the expected number of errors once it has learnt the
rules of the game. There are many situations were it is
reasonable for an agent or predictor to have some grace or
learning period.

Formally, for a given sequence $x_{1:n}$ and the Hellinger
distance $h_t$ defined in Subsection~\ref{sec:conv} between
$M$ and the true environment $\mu$, the future expected error
is written as follows:
\[
  \sum_{t=n+1}^\infty\E[h_t|x_{1:n}]
\]
In general, there is no way of knowing how similar the future
will be to the past. The environment may give the agent a long
simple sequence of observations which it predicts well and then
suddenly change to a complex unpredictable pattern. Here it is
worth mentioning the relation to what Hume called the principle
of uniformity of nature \cite{Hume:1739}. He argued that
induction implicitly assumes this principle which states that
the future will resemble the past. But the only way to verify
this principle is through induction which is circular reasoning
and therefore induction cannot be proven.
As we will see however, the future bound does not assume this
principle but rather accounts for it through a factor that
measures the divergence of the future from the past.

In the total bound ($n=0$), a sudden switch to a more complex
sequence is accounted for by the Kolmogorov complexity of the
true environment $K(\mu)$. For these surprising complex
environments, $K(\mu)$ is obviously large and hence the bound
is weaker which automatically accounts for errors that will be
made after the environment switches. A finite total number of
errors only implies that the number of future errors {\em
eventually} converges to zero. The purpose of the following
future bound however is to use the information obtained so far
in $x_{1:n}$ to give a tighter, more useful bound.

In the case of future errors, the bound is again in terms of
the Kolmogorov complexity of the true environment but now it is
conditioned on the past observation sequence $x_{1:n}$.
Formally it can be shown \cite{Hutter:07postbndx} that
\[
  \sum_{t=n+1}^\infty \E[h_t|x_{1:n}] \;\leq\; (K(\mu|x_{1:n})+K(n))\ln 2 + O(1)
\]
This bound is highly intuitive. The dominant term,
$K(\mu|x_{1:n})$, depends on the relationship between what has
been seen so far and what will be seen in the future. As
described previously, this Kolmogorov complexity measures the
complexity of environment $\mu$ in relation to side information
$x_{1:n}$. Therefore if the principle of uniformity holds and
the future does resemble the past, then this term is small. This
means that once a sample has been seen which is large enough to
be representative of the entire population or the future
sequence, then the future number of prediction errors is
expected to be relatively small. On the other hand if the
future is going to be vastly different from the past then this
history $x_{1:n}$ is not much help at all for understanding
$\mu$, so the future bound $K(\mu|x_{1:n})$ will be almost as
high as the initial total bound $K(\mu)$. In other words, the
bound measures the value of the history $x_{1:n}$ for learning
the future.

Consider for instance a sequence of identical images $I$ of
high complexity $K(I)\approx 10^6$. Since the true environment
$\mu$ simply repeats this image it also has complexity
$K(\mu)\approx 10^6$. After making even just one observation of
$I$ the dramatic difference between the total bound and the
total future bound becomes apparent. The total bound on the
number of errors is proportional to the complexity of the true
environment $K(\mu)$, so all this tells us is that the total
number of errors will be approximately one less than $10^6$.
But since this sequence is simply a repetition of the image $I$
we expect a much smaller number of errors. The bound above
tells us that the total number of future errors is bounded by
$(K(\mu|I)+K(1))\ln2$ after seeing $I$ once. But the complexity
of $\mu$ in relation to $I$ is $O(1)$ since $\mu$ only needs to
copy and paste $I$ repeatedly. Therefore the total number of
future errors will be very small which agrees with intuition.
This example illustrates the potential value of this future
bound.

\subsection{Universal is Better than Continuous $\M$ }

It has been argued that the universal class and prior are an
excellent choice for completing the Bayesian framework as they
lead to Solomonoff's $M=\xi_U$, which we have shown to be
(Pareto) optimal and to perform excellently on any computable
deterministic or stochastic environment $\mu$. Although this
assumption of computability is reasonable, the question of its
performance in incomputable environments remains interesting.
For example, we may want to know how it compares with other
methods in predicting a biased coin being flipped where the
bias $\t$ is an incomputable number. Intuitively this should
not be a major problem as the computable numbers are dense in
the reals and therefore there are computable hypotheses $\t'$
that are arbitrarily close to $\t$.

It is indeed possible to formally prove that $M$ does as good
as any Bayesian mixture $\xi$ over any model class $\M$,
continuous or discrete, and prior function $w()$ over this
class. The reason for this is that although a specific
environment $\nu$ in $\M$ may be incomputable and its prior
$w_\nu$ may be zero, the prior function $w()$ and the overall
mixture $\xi$ generally remain computable. This is precisely
what occurs in Laplace's rule for i.i.d.\ data. In this case
the model class is $\M=\{\t : \t\in[0,1]\}$ and the prior density
$w(\t)=1$ for any $\t$. The majority of the values of $\t\in\M$
are incomputable since $\M$ is uncountable. Since $w()$ is a
density it assigns a prior of zero to any particular value, but
for any partial hypothesis $\t_p=(a,b)\subseteq [0,1]$ where
$b-a$ is incomputable, $w(\t_p)$ is also incomputable. But the
function $w()$ is itself computable, in fact very simple, and
the Bayesian mixture $\xi(x)=\frac{s!f!}{(s+f+1)!}$ (for $s$
successes and $f$ failures in $x$) is also computable and quite
simple. This computability of $\xi$ implies the following
general result for any, possibly incomputable, environment
$\mu$:
\[
  KL_n(\mu||M) \;<\; KL_n(\mu||\xi)+K(\xi)\ln2 + O(1)
\]
The Kullback-Leibler divergence
$KL_n(\mu||\rho):=\E[\ln(\mu(x_{1:n})/\rho(x_{1:n}))]$
itself upper bounds the various total losses defined in
Section~\ref{cha:bayes}. For here it suffices to know that
small $KL$ implies small total error/loss. If $\xi$ is
computable, which is usually the case even for continuous
classes, then $K(\xi)\ln2$ is finite and often quite small.
The importance of this result is that it implies that classical
Bayesian bounds for $\xi$ based on continuous classes like the
one in Subsection~\ref{sec:cont} also hold for the universal
predictor $M$ even for the incomputable $\mu$ in continuous
classes $\M$. This means that $M$ performs not much worse than any
Bayesian mixture $\xi$ (and often much better). Indeed, even
more general, if there exists any computable predictor that
converges to $\mu$ in KL-divergence, then so does $M$, whatever
$\mu$.
This means that if there is any computable predictor that
performs well in some incomputable environment $\mu$, then $M$
will also perform well in this environment.

\section{Approximations and Applications}\label{cha:app}

Solomonoff's universal probability $M$ as well as Kolmogorov
complexity $K$ are not computable, hence need to be
approximated in practice.
But this does not diminish at all their usefulness as gold
standards for measuring randomness and information, and for
induction, prediction, and compression systems.
Levin complexity $Kt$ is a down-scaled computable variant of
$K$ with nice theoretical properties, the context tree
weighting (CTW) distribution may be regarded as a very
efficient practical approximation of $M$, and the minimum
encoding length principles (MDL\&MML) are effective model
selection principles based on Ockham's razor quantifying
complexity using practical compressors.
$K$ and $M$ have also been used to well-define the clustering
and the AI problem, and approximations thereof have been
successfully applied to a plethora of complex real-world
problems.

\subsection{Golden Standard}

\begin{quote}
{\em ``in spite of its incomputability,
Algorithmic Probability can serve as a kind of `Gold standard' for
inductive systems''} - Ray Solomonoff, 1997.
\end{quote}

It is useful to keep this above attitude in mind when looking
at how Solomonoff induction can be applied. This idea of a
`gold standard' is certainly not new. There are various
problems or scientific areas where an `optimal' solution or
model is known but for practical purposes it must be
approximated.

A simple example of this concept can be seen in the problem of
playing optimal chess. It is known that a min-max tree search
approach extended all the way to the end of a game will yield
optimal play, but this is entirely infeasible due to the huge
branching factor and extremely large depth of the tree. The
magnitude of the tree required to solve chess is truly massive,
with more possible chess games than there are atoms in
the universe. Instead this approach is approximated using much
smaller trees and heuristics for pruning the tree and
evaluating board positions. These approximations can however
perform very well in practice which was comprehensively
demonstrated in 1997 when the chess computer Deep Blue finally
defeated the reigning world Champion Gary Kasparov.

Even given the exponentially increasing processing power of
computers there is no foreseeable future in which the
computation of a complete chess min-max tree will be feasible.
An infeasible problem is a problem that can not be practically
computed and therefore must be approximated. An incomputable
problem such as Solomonoff induction can also not be
practically computed but can only be asymptotically computed
given infinite resources. But when looking at practical
applications we only consider feasible approximations and
therefore it may be argued that there is little difference
between an incomputable problem and an entirely infeasible
problem. Rather it is important to find approximations that are
useful, although this may often be more likely for problems
that are only infeasible rather than incomputable.

Another example of a `golden standard' is quantum
electrodynamics in relation to chemistry and physics. Quantum
electrodynamics was called the `jewel of physics' by Richard
Feynman because of its ability to make extremely accurate
predictions of various phenomena. Theoretically this theory is
able to accurately predict essentially any atomic interactions
which of course extends to all chemical processes. However
applying this theory to almost any non-trivial interactions
requires extremely complex or infeasible calculations. Because
of this, other methods are generally used which are, or can be
considered, approximations of quantum electrodynamics, like the
extremely successful Hartree-Fock approximation.

Even competitive Rubik's cube solving may be seen as human
approximation of a known, and computable, gold standard. It has
been shown that a standard Rubik's cube can be solved, from any
position, in 20 moves. This is the theoretically best that can
be achieved but any human attempt is still far from this. The
level of planning and visualization necessary to compute this
optimal solution is infeasible for a human but it does give a
lower bound or `gold standard' on what is possible.

Even if a `gold standard' idea or theory cannot be implemented
in practice, it is still valuable to the concerned field of
study. By demonstrating a theoretically optimal solution to
universal induction, Solomonoff is useful both as a guide for
practical approximations and as a conceptual tool in
understanding induction. By adding theoretically optimal
decision making to this framework, the second author shows in
\cite{Hutter:04uaibook} that the resulting AIXI model can
similarly be considered a `gold standard' for Artificial
Intelligence.

Just as in the examples above, Solomonoff induction and
Kolmogorov complexity can and have been successfully
approximated.

\subsection{Minimum Description Length Principle}

The Minimum Description Length (MDL) principle is one such
approximation of Solomonoff induction which is based on the
idea that inductive inference can be achieved by finding the
best compression of data since this requires learning and
understanding the underlying structure of the data. Put simply
it predicts data that minimizes the compression of this new
data with the old data. MDL achieves computational feasibility
by restricting the hypotheses, which are the methods of data
compression, to probabilistic Shannon-Fano based encoding
schemes \cite{Gruenwald:07book}. This principle is actually
based on a similar earlier concept known as the minimal message
length principle which considers a message to consist of the
size of a model plus the size of the compressed data using this
model \cite{Wallace:05}.

\subsection{Resource Bounded Complexity and Prior}

Levin Complexity is a direct variant of algorithmic complexity
which is computable because it bounds the resources available
to the execution. Specifically it uses a time-bounded version
of Kolmogorov complexity defined as follows:
\[
  Kt(x) \;=\; \min_p\{\length(p)+\log(time(p))\,:\,U(p)=x\}
\]
This avoids the problem of short programs $p$ that run
indefinitely and hence may or may not produce $x$. Eventually
the time factor will invalidate them as candidates and hence
they can be terminated \cite{Li:08}. The speed prior $S(x)$
similarly approximates $M(x)$ by limiting the time available
\cite{Schmidhuber:02speed}.

Even off-the-shelf compressors such as Lempel-Ziv
\cite{Lempel:76} can be considered approximations of Kolmogorov
complexity to varying degrees. The more advanced the techniques
used to extract the structure from and hence compress a file,
the better the approximation. In fact, one of the key insights
by Solomonoff was that induction and compression are
essentially equivalent problems.

The Kolmogorov complexity is an extremely broad concept which
applies to any effectively describable object. In practice
however we often only need to measure the complexity of an
object relative to a specific class of objects, and this can
make approximations much easier.

\subsection{Context Tree Weighting}

For example in the Context Tree Weighting (CTW) algorithm,
described below, we want to find the complexity of a specific
suffix tree measured against the class of all prediction suffix
trees. Using a simple encoding scheme it is easy to create a
set of binary prefix-free strings that are in one to one
correspondence with the set of prediction suffix trees. These
are therefore short unambiguous descriptions of each
tree/environment and can be used to give an effective
approximation of Kolmogorov complexity. If $z$ is the
prefix-free string that describes the prediction suffix tree
$p$, the Kolmogorov complexity of $p$ is approximated by
$\length(z)$. This approximation is also intuitive, since
simpler trees have shorter descriptions and the corresponding
prior $2^{-\length(z)}$ when summed over all trees gives 1.

The context tree weighting algorithm, originally developed as a
method for data compression \cite{Willems:95}, can be used to
approximate Solomonoff's universal prior $M(x)$. It is worth
particular mention due to its elegant design and efficient
performance. Universal induction is incomputable both because
of the Kolmogorov complexity necessary to establish the priors
and because of the infinite sum over Turing machines. It is
often possible to establish an intuitively satisfying
approximation of an environment's complexity, as mentioned
above with prediction suffix trees. However it is much more
difficult to think of an efficient method of summing over a
large, or infinite, class of environments. It is this property
that makes the CTW algorithm noteworthy.

The CTW algorithm considers the class of all prediction suffix
trees up to some fixed depth $d$ (it can also be implemented to
use arbitrarily deep trees). The number of these grows double
exponentially with the depth $d$, so this quickly becomes a
very large class. Remarkably the mixture model over this entire
class, weighted according to the Kolmogorov complexity
approximation above, turns out to be exactly equal to the
probability given by a single structure called a context tree.
This context tree can be implemented and updated very
efficiently and therefore leads to a computationally feasible
and effective approximation. Although this class is massively
reduced from the universal class of all computable
distributions, it is sufficiently rich to achieve excellent
compression and has been shown to perform well in many
interesting environments \cite{Hutter:11aixictwx}.

\subsection{Universal Similarity Measure}

One interesting application of Kolmogorov complexity is in the
constructions of a universal similarity measure. Intuitively
two objects can be considered similar if it takes little effort
to transform one into the other. This is precisely the concept
captured using conditional Kolmogorov complexity since it
measures the complexity of one string given another string.
Formally, for objects $x$ and $y$, the similarity metric is
defined by symmetrizing and normalizing $K(x|y)$ as follows:
\[
  S(x,y) \;:=\; \frac{\max\{K(x|y),\, K(y|x)\}}{\max\{K(x),\, K(y)\}}
\]
Since it takes no effort to change an object into itself it is
clear that $K(x|x)\approx 0$ and hence $S(x|x)\approx 0$. If
there is no similarity between $x$ and $y$ then $K(x|y)\approx
K(x)$ and $K(y|x) \approx K(y)$ and therefore $S(x,y) \approx
K(x)/K(x) = K(y)/K(y) = 1$. Although the Kolmogorov complexity is
incomputable, it is possible to achieve an effective approximation
of this metric by approximating $K$ with a good compressor such
as Lempel-Ziv, gzip, bzip, or CTW.

When comparing many objects, a similarity matrix is created by
comparing each object to each other using the above
metric. This matrix can then be used to construct trees based
on the similarity of objects. This method has been used to
achieve some astounding results \cite{Li:04}. For example, it
was used to accurately construct the exact evolutionary tree of
24 mammals based only on the structure of their DNA. It was
also used to construct a tree of over 50 languages showing the
common roots and evolution of these languages. Other successes
include correctly categorizing music by their composers and
files by their type \cite{Cilibrasi:05}.

\subsection{Universal Artificial Intelligence}

Universal artificial intelligence involves the design of agents
like AIXI that are able to learn and act rationally in
arbitrary unknown environments. The problem of acting
rationally in a known environment has been solved by sequential
decision theory using the Bellman equations. Since the unknown
environment can be approximated using Solomonoff induction,
decision theory can be used to act optimally according to this
approximation. The idea is that acting optimally according to
an optimal approximation will yield an agent that will perform
as well as possible in any environment with no prior knowledge.

This unification of universal induction and decision theory
results in the AIXI agent \cite{Hutter:04uaibook}. According to
Legg and Hutter's definition of intelligence, which they argue
to be reasonable \cite{Hutter:07iorx}, this AIXI agent is the
most intelligent agent possible. Since this is an extension of
Solomonoff induction it is clearly also incomputable, but again
practical approximations exist and have produced promising
results.

The MC-AIXI-CTW approximation builds on the context tree
weighting algorithm described above by incorporating actions
and rewards into the history and using an efficient Monte-Carlo
tree search to make good decisions. Just as the context tree
weighting algorithm is an efficient approximation of
Solomonoff, MC-AIXI-CTW is an efficient approximation of AIXI
\cite{Hutter:11aixictwx}.

\section{Discussion}\label{cha:discussion}

Solomonoff \cite{Solomonoff:64} created a completely general
theory of inductive inference. Subsequent developments
\cite{Hutter:07uspx} have shown that his system solves many of
the philosophical and statistical problems that plague other
approaches to induction.
In the same/similar sense as classical logic solves the problem
of how to reason deductively, Solomonoff solved the problem of
how to reason inductively.
No theory, be it universal induction or logic or else, is ever
perfect, but lacking better alternatives, over time people
often get used to a theory, and ultimately it becomes a
de-facto standard. Classical logic is there already; Bayesian
statistics and universal induction are slowly getting there.
In this final section we briefly discuss the major pro's and
con's of Solomonoff induction before concluding.

\subsection{Prior Knowledge}

Philosophically speaking our predictions of the future are
dependent on a lifetime of observations. In practice we need
only consider relevant information to make useful prediction
but this question of relevance is not always straightforward.

One of the problems with Solomonoff induction is that relevant
background knowledge is not explicitly accounted for. There are
two ways to modify Solomonoff induction to account for prior
background knowledge $y$.

The first method is to judge the complexity of each environment
$\nu$ in relation to this prior knowledge $y$. Formally this is
done by providing $y$ as extra input to the universal Turing
machine when computing $K(\nu)$ which leads to the conditional
Kolmogorov complexity $K(\nu|y)$. One of the properties of the
conditional Kolmogorov complexity $K(a|b)$ is that it can be no
bigger than the Kolmogorov complexity $K(a)$ alone. This is
because the extra input $b$ can simply be ignored. This means
that all environments, in particular the true environment, are
weighted at least as high as they would be without the prior
knowledge $y$. By biasing the prior towards environments that
can be more easily expressed using our prior knowledge this
method should intuitively never lead to worse convergence and
often to faster convergence.

The second method is to prefix the observation sequence $x$,
which we wish to predict, with the prior knowledge $y$. We are
therefore predicting the continuation of the sequence $yx$.
This method treats background knowledge as previous
observations. In the case of an ideal agent that applies
Solomonoff's method as soon as it is initialized, or from
`birth', all previous knowledge is always contained in its
history, so this method of prefixing $x$ with $y$ occurs
implicitly.

In practice, the best method may depend on the type of
background knowledge $y$ that is available. If $y$ is in the
form of previous observations, perhaps taken from similar
experiments or situations in the past, then it would be natural
to prefix $x$ with $y$. On the other hand, if $y$ is a partial
hypothesis which we know, or strongly suspect to be correct, it
would be more appropriate to judge the complexity with the
conditional Kolmogorov complexity given $y$. This can be
thought of as looking at the complexity of different methods of
completing the partial model we currently have. Although these
different methods are conceptually useful due to different
possible types of background knowledge, they are theoretically
nearly equivalent. Intuitively this is because in both cases
any information present in the background knowledge $y$ is
being used to simplify the model for $x$ and it may not matter
if this knowledge is provided with $x$ or separately.

\subsection{Dependence on Universal Turing Machine $U$}

A common criticism of both Solomonoff's universal predictor and
the Kolmogorov complexity is the dependence on the choice of
universal reference machine $U$. As we discussed in
Section~\ref{sec:kolmo}, the major problem is the difficulty of
agreeing on a formal definition of a `natural' Turing machine.
We argued that this is equivalent to agreeing on the definition
of a natural encoding scheme which may seem simpler intuitively
however the core problem remains. In any case, the choice of
$U$ is only an issue for `short' sequences $x$. This is because
for short $x$ and any arbitrary finite continuation $y$ we can
always choose $U$ that predicts $y$ to follow $x$. However as
the length of $x$ increases, the predicted continuation of $x$
will become independent of $U$.

It is worth noting that this problem of arbitrary predictions
for short sequences $x$ is largely mitigated if we use the
method of prefixing $x$ with prior knowledge $y$. The more
prior knowledge $y$ that is encoded, the more effective this
method becomes. Taken to the extreme we could let $y$ represent
all prior (scientific) knowledge, which is possibly all
relevant knowledge. This means that for any $x$ the string $yx$
will be long and therefore prediction will be mostly unaffected
by the choice of universal reference machine.

\subsection{Advantages and Disadvantages}

As we have seen, the universal Solomonoff approach has many
favorable properties compared to previous methods of induction.
In particular these include:
\begin{itemize}\parskip=0ex\parsep=0ex\itemsep=0ex
\item General total bounds for generic class, prior and
    loss function as well as instantaneous and future
    bounds for both the i.i.d.\ and universal cases.
\item The bound for continuous classes and the more general
    result that $M$ works well even in non-computable
    environments.
\item Solomonoff satisfies both reparametrization and
    regrouping invariance.
\item Solomonoff solves many persistent philosophical
    problems such as the zero prior and confirmation
    problem for universal hypotheses. It also deals with
    the problem of old evidence and we argue that it
    should solve the black ravens paradox.
\item The issue of incorporating prior knowledge is also
    elegantly dealt with by providing two methods which
    theoretically allow any knowledge with any degree of
    relevance to be most effectively exploited.
\end{itemize}
These significant advantages do however come at a cost.
Although it seems Solomonoff provides optimal prediction given
no problem-specific knowledge, the persistent multiplicative or
additive constants in most results mean this optimality may
only occur asymptotically. For short sequences and specific
problems, Solomonoff may not perform as well as other methods,
but an automatic remedy is to take into account previous or
common knowledge. Of course the most significant drawback to
this approach is the incomputability of the universal
predictors $M$ and $\xi_U$. Although approximations do exist,
with some very interesting results, they are still either
infeasible or fairly crude or limited approximations.

\subsection{Conclusion}\label{sec:conclusion}

In this article we have argued that Solomonoff induction solves
the problem of inductive inference. Not only does it perform
well theoretically and solves many problems that have plagued
other attempts, it also follows naturally from much of the
significant work in induction that precedes it.

Initially we explained how the subjectivist interpretation of
probability is appropriate for dealing with induction since
induction concerns our beliefs about the unknown. Cox's
demonstration that the standard axioms of probability are a
necessary consequence of any rational belief system provides
strong theoretical justification for using the standard axioms
while maintaining a subjectivist approach.

Bayes theorem is derived directly from these standard
probability axioms and, when combined with the reasonable setup
of holding some class of possible explanations, each with some
prior plausibility, we arrive at the Bayesian framework. This
framework provides a rational method of updating beliefs and in
some sense it provides the optimal method of predicting future
observations based on current beliefs and evidence. We argued
that the concept of prediction is general enough to encompass
all relevant problems of inductive inference and therefore this
Bayesian framework is also optimal for inductive inference.

This framework requires that we have some pre-conceived beliefs
about the possible causes for our observations. Ideally these
beliefs are formed before any evidence has been observed so
they must be based on an objective and universally acceptable
principle. By combining the ideas of Occam and Epicurus we are
at such a principle which is both intuitive and philosophically
appealing and applies to the largest reasonable class of
possible environments. A strict formalization of this principle
is achieved through contributions of Turing and Kolmogorov.

Completing the Bayesian framework with this universal model
class and prior produces Solomonoff induction. In a very loose
sense this demonstrates that Solomonoff induction follows as a
natural consequence of rationality. Although incomputable, this
universal method for inductive inference can be used to guide
practical implementations and also gives us new insights into the
mechanics of our own thinking.

By refuting Maher's invalid claim that probability alone can
solve the problem of confirmation, we have shown that these
failings of other systems of induction remain present, while
Solomonoff deals with confirmation successfully by assigning
non-zero priors to all computable hypotheses. We have also
looked briefly at how Solomonoff induction deals with the black
ravens paradox and argued that it should give the desired result.

We hope that this accessible overview of Solomonoff induction
in contemporary and historical context will bring an
appreciation and awareness of these ideas to a larger audience,
and facilitates progress in and applications of this exciting
and important field.


\addcontentsline{toc}{section}{\refname}

\begin{small}
\newcommand{\etalchar}[1]{$^{#1}$}

\end{small}

\end{document}